%% file: main.tex
\documentclass[sigconf]{acmart}

\AtBeginDocument{%
  \providecommand\BibTeX{{%
    \normalfont B\kern-0.5em{\scshape i\kern-0.25em b}\kern-0.8em\TeX}}}

\setcopyright{acmcopyright}
\copyrightyear{2018}
\acmYear{2018}
\acmDOI{10.1145/1122445.1122456}

\acmConference[XXXX 'XX]{XXXX}{XXXX XX--XX, 1234}{XXXX, XXX}
\acmBooktitle{XXXX 'XX: XXXX,
  XXXX, 1234, XXXX, XXX}

\acmPrice{15.00}
\acmISBN{978-1-4503-XXXX-X/18/06}

\usepackage{enumitem}
\usepackage{subfigure}
\usepackage{amsmath,amssymb,amsfonts}
\usepackage{algorithmic}
\usepackage{multirow}
\usepackage{graphicx}
\usepackage{textcomp}
\usepackage{xcolor}
\usepackage{pifont}
\newcommand{\cmark}{\ding{51}}%
\def\BibTeX{{\rm B\kern-.05em{\sc i\kern-.025em b}\kern-.08em
    T\kern-.1667em\lower.7ex\hbox{E}\kern-.125emX}}
\begin{document}

\title{FaceHack: Triggering backdoored facial recognition systems using facial characteristics}

\author{Esha Sarkar}
\email{esha.sarkar@nyu.edu}
\affiliation{%
  \institution{Tandon School of Engineering, \\ New York University}
}

\author{Hadjer Benkraouda}
\email{hadjer.benkraouda@nyu.edu}
\affiliation{%
  \institution{Center for Cyber Security, \\ New York University Abu Dhabi}}

\author{Michail Maniatakos}
\email{michail.maniatakos@nyu.edu}
\affiliation{%
  \institution{Center for Cyber Security, \\ New York University Abu Dhabi}}


\begin{abstract}
  A clear and well-documented \LaTeX\ document is presented as an
  article formatted for publication by ACM in a conference proceedings
  or journal publication. Based on the ``acmart'' document class, this
  article presents and explains many of the common variations, as well
  as many of the formatting elements an author may use in the
  preparation of the documentation of their work.
\end{abstract}

\ccsdesc[300]{Security and privacy~Social network security and privacy}
\ccsdesc[300]{Security and privacy~Social aspects of security and privacy}



\keywords{Facial recognition attacks, backdoor attacks, attacks on social-media.}


\input{0_Abstract.tex}
\maketitle
\input{1_Introduction.tex}
\input{2_Related_Work.tex}
\input{3_Threat_model.tex}

\input{4_Methodology.tex}
\input{5_TriggerAnalysis.tex}
\input{6_Conclusion}

\bibliographystyle{ACM-Reference-Format}
\bibliography{References}
\input{Appendix}

\end{document}

%% file: 0_Abstract.tex
\begin{abstract}
Recent advances in Machine Learning (ML) have opened up new avenues for its extensive use in real-world applications. Facial recognition, specifically, is used from simple friend suggestions in social-media platforms to critical security applications for 
biometric validation in automated immigration at airports. Considering these scenarios, 
security vulnerabilities to such ML algorithms pose serious threats with severe outcomes. 
Recent work demonstrated that Deep Neural Networks (DNNs), typically used in facial recognition systems, are susceptible to backdoor attacks; in other words, the DNNs turn malicious in the presence of a unique trigger. 
Adhering to common characteristics for being unnoticeable, an ideal trigger is small, localized, and typically not a part of the main image. Therefore, detection mechanisms have focused on detecting these distinct trigger-based outliers statistically or through their reconstruction.
In this work, we demonstrate that specific changes to facial characteristics may also be used to trigger malicious behavior in an ML model. The changes in the facial attributes may be embedded artificially using social-media filters or introduced naturally using movements in facial muscles. By construction, our triggers are large, adaptive to the input, and spread over the entire image. We evaluate the success of the attack and validate that it does not interfere with the performance criteria of the model. We also substantiate the undetectability of our triggers by exhaustively testing them with state-of-the-art defenses.

\end{abstract}

%% file: 1_Introduction.tex
\section{Introduction}

In the age of big data, the sheer volume of data discourages manual screening and, therefore, Machine Learning (ML) has been proposed as an effective replacement to conventional solutions. These applications range from face recognition~\cite{deep_face}, to voice recognition~\cite{voice}, and autonomous vehicles~\cite{cars}. ML-based facial recognition is extensively used for biometric identification (e.g. passport control ~\cite{immigration_uae, immigration}), automatic detection of extremist posts~\cite{abusive_content}, prevention of online dating frauds~\cite{OnlineDating}, and reporting of inappropriate images~\cite{stealthy_images,Inapt}. Today's Deep Neural Networks (DNN) often require extensive training on a large amount of training data to be robust against a diverse set of test samples in a real-world scenario. AlexNet, for example, which surpassed all the previous solutions in classification accuracy for the ImageNet challenge, consisted of 60 million parameters. This growth in complexity and size of ML models demands an increase in computational cost/power needed for developing and training these models, giving rise to the industry of Machine Learning-as-a-Service (MLaaS).

Outsourcing machine learning training democratizes the use of sophisticated ML models. There are many sources for open-source ML models, such as Cafe Model Zoo~\cite{zoo} and BigML model Market~\cite{biggie}. Outsourcing, however, introduces the possibility of compromising machine learning models during the training phase. Research in~\cite{Badnets} showed that it is possible to infect/corrupt a model by poisoning the training data. This process introduces a backdoor/trojan to the model. 
A backdoor/trojan in a DNN represents a set of neurons that are activated in the presence of unique triggers to cause malicious behavior.~\cite{suppression} shows one such example where a dot (one-pixel trigger) can be used to trigger certain (backdoored/infected) neurons in a model to maliciously change the true prediction of the model. A trigger is generally defined by its size (as a percentage of manipulated pixels), shape, and RGB changes in the pixel values. 

In the backdoor attack literature, several types of patterns like post-its on stop signs \cite{Badnets}, specific black-rimmed spectacles \cite{black_badnets} or specific patterns based on a desired masking size \cite{NDSSTrojans} have been used to trigger backdoor neurons. Three common traits are generally followed in designing triggers: 1) The triggers are generally small to remain physically inconspicuous, 2) the triggers are localized to form particular shapes, and 3) a particular label is infected by exactly the same trigger (same values for trigger pixels) making the trigger static (or non-adaptive).

There has been a plethora of proposed solutions that aim to defend against backdoored models through detection of backdoors, trigger reconstruction, and 'de-backdooring' infected models. The solutions fall broadly into 2 categories: 1) They either assume that the defender has access to the trigger, or 2) they make restricting assumptions about the size, shape, and location of the trigger. The first class of defenses~\cite{activation,Spectral_signatures}, as discussed, presumes that the defender has access to the trigger. Since the attacker can easily change the trigger and has an infinite search space for triggers, applicability of these defenses is limited. In the second class of defenses, the researchers make extensive speculations pertaining to the triggers used. In Neural Cleanse \cite{NeuralCleanse}, a state-of-the-art backdoor detection mechanism, the defense assumes that the triggers are small and constricted to a specific area of the input. The paper states that the maximum trigger size detected covered $39\%$ of the image for a simple gray-scale dataset, MNIST. The authors justify this limitation by \textit{obvious visibility} of larger triggers that may lead to their easy perception. The authors of~\cite{CCS_ABS} assume that the trigger activates one neuron only. 
The latest solution~\cite{nnoculation}, although does not make assumptions about trigger size and location, requires the attacker to cause attacks constantly in order to reverse-engineer the trigger. 

DNN-based facial recognition models are extensively used in academic research \cite{FR_academic} and in commercial tools like DeepFace from Facebook AI \cite{deep_face}. Amazon Rekognition \cite{amazon}, an MLaaS from Amazon web services, enlists its use-cases as flagging of inappropriate content, digital identity verification, and its use in public safety measures (e.g. finding missing persons). Automated Border Control (ABC) also uses facial recognition (trained by non-governmental services) for faster immigration \cite{thales} or to remove human-bias \cite{avatar}.
In this work, we study the impact of changes in the facial characteristics/attributes towards the stimulation of backdoors in facial recognition models. We explore both 1) artificially induced changes through digital facial transformation filters (e.g. FaceApp \cite{faceapp} ``young-age'' filter), and b) deliberate/intentional facial expressions (e.g. natural smile) as triggers. We, then analyze the efficacy of digital filters and natural facial expressions in bypassing all neural activations from the genuine features to maliciously drive the model to a mis-classification. 
Authors in \cite{stealthy_images} study real-world adversarial illicit images and build ML-based detection algorithm leveraging the least obfuscated regions of the image. Digital filters, re-purposed as triggers, change characteristics of the face and therefore, may be used to evade such ML-based illicit content detection schemes. Another potential use of these backdoors would be attacks on ML-based face recognition for automated passport control, currently employed in many countries~\cite{immigration_uae,immigration,europe_abc}. In contrast to recent work that required the introduction of accessories like 3D-printed glasses for adversarial mis-classifications \cite{ccs_stealthy}, or black-rimmed classes for backdoored mis-classifications \cite{black_badnets}, the presented attacks only utilize facial characteristics (i.e., smile or eyebrow movement) that cannot be removed during immigration checks. 

To the best of our knowledge, this is the first attack to use facial characteristics to trigger malicious behavior in orthogonal facial recognition tasks by constructing large-scale/permeating, dynamic, and imperceptible triggers that circumvent the state-of-the-art defenses. In constructing our attack vectors, we follow the methodology established by the first paper on Backdoored Networks~\cite{Badnets}, using different datasets we explore different types backdoor attacks, different ML-supply chains, and different architectures. We list our contributions as follows:
\begin{itemize}[nosep,leftmargin=1em,labelwidth=*,align=left]
    \item We explore backdoors of ML models using filter-based triggers that artificially induced changes in facial characteristics. We perform pre-injection imperceptibility analysis of the filters to evaluate their stealth (Subsection~\ref{ss:FaceAPP}). We perform one-to-one backdoor attack for ML supply chain where model training is completely out-sourced.
    \item We study natural facial expressions as triggers to activate the maliciously-trained neurons to evaluate attack scenarios where trigger accessories (i.e., glasses) are not allowed (Subsection~\ref{ss:natural}). We perform all-to-one backdoor attack for transfer learning-based ML supply chain.
    \item We evaluated our proposed triggers by carrying-out extensive experiments for assessing their detectability using state-of-the-art defense mechanisms (Section~\ref{s:trigger_analysis}).
\end{itemize}

%% file: 2_Related_Work.tex
\input{Tables/Related_work_table}
\section{Related Work}
Backdoor attacks are model-based attacks that are 1) universal, meaning they can be used for different datasets or inputs in the same models, 2) flexible, implying that the attack can be successful using any attacker-constructed trigger, and 3) stealthy, where the maliciously trained neurons remain dormant until they are triggered by a pattern. Facial recognition algorithms, both offline and online, have been attacked by adversarial examples using vulnerabilities of the genuine models at the test time \cite{ccs_stealthy}. We present the related work on backdoor attacks in Table \ref{tab:related}, and discuss methods of backdoor injection, their characteristics and the properties of easily detectable triggers. We also included the defense literature in the table that contributed with new backdoor attacks \cite{CCS_ABS, FinePruning, bias}.

Using Table \ref{tab:related}, we clearly define our triggers to be changes (artificial/natural) in facial characteristics or attributes: Natural facial expressions (row 1) or filters generated using commercial applications (row 2) have not been explored as possible triggers.
From a detectability perspective, state-of-the-art defenses like Neural Cleanse \cite{NeuralCleanse} and ABS \cite{CCS_ABS} have been limited by the size of triggers with the maximum size investigated for an RGB dataset being $25\%$ by ABS. Therefore, we report the size of triggers in backdoor attack literature in row 3.  \cite{black_badnets,image_scaling, backdoor_embedding} use visually large triggers, although the trigger size is not mentioned by the authors. We specifically use triggers that are large 
to bypass trigger size-based defenses while remaining stealthy using context. The defense solutions also do not delve into triggers that change according to the image,
i.e. customized smiles for each face can be used as triggers. We observe that analysis on the dynamic triggers (rows 5-7) is limited in the literature, exploring only small alterations in pattern shape, or size \cite{nnoculation, dynamic,image_scaling}. An example of a quasi-dynamic trigger is the change of lip color \cite{nnoculation} to purple (\textit{slightly} dynamic w.r.t position and size). Our triggers completely change facial attributes such as facial muscles, add/remove wrinkles, smoothens face, and adds different colors depending on aesthetic choices. Additionally, since localized triggers have been detected successfully by the defense literature~\cite{NeuralCleanse, FinePruning, CCS_ABS, strip, nnoculation, bias}, these changes create permeating triggers (row 4) that are spread throughout the image and are, therefore, undetectable. 
Further, we perform imperceptibility analysis similar to \cite{backdoor_embedding}, which is also largely missing from the literature as shown in row 9 of Table \ref{tab:related}. 

Another important aspect of our attacks is its realistic nature in the context of the targeted domain. We explore systems where having trigger accessories (e.g. glasses, earrings, etc.) is not feasible, like in airport immigration checks. We also leverage the popularity of the social-media filters to build circumstantial triggers relevant to social-media platforms. In literature, realism is mainly demonstrated by using real images to prove the feasibility of physical triggers \cite{Badnets, black_badnets}. Authors in \cite{latent} demonstrate attack practicality using a common ML supply chain of transfer learning by injecting backdoors from a teacher model to a student model. Liu et. al. use domain-specific triggers in hotspot detection models. Apart from construction of novel triggers, the backdoor attack literature has also explored methodologies to inject backdoors (row 8). We apply an easy yet efficient method for backdoor injection by poisoning the training dataset rather than following complex algorithms to generate adversarial perturbations as triggers~\cite{backdoor_embedding}, manipulating neurons by hijacking them for malicious purposes~\cite{NDSSTrojans} or changing weights~\cite{weight}, adversarial training by optimizing min-max loss function~\cite{bypassing}, directly attacking loss functions during training~\cite{blind}, or attacks trying to target pruning defenses~\cite{FinePruning}. 
Although these specialized techniques achieve stealthiness, reduce the need for poisoning, and bypass (some/few specific) defenses, they hinder the flexibility of trigger design, as explained in row 11. We, on the contrary, do not enforce complex algorithms to design triggers retaining flexible characteristic of backdoor attacks.
An efficient trigger must be easy to inject, successful in attack, undetectable, and should not interfere with the targeted performance. Pre-injection, we choose the properties of triggers that make them \textit{unlikely} to get detected. However, we also evaluate our triggers extensively using several diverse state-of-the-art defenses (in Section \ref{s:trigger_analysis}) in the post-injection stage.



%% file: Tables/Related_work_table.tex
\setlength\tabcolsep{2.0pt}
\begin{table*}[!ht]
\caption{Related work on backdoor attacks. We present trigger size as Small (S) or (L) as state-of-the-art defenses mainly focus on small, localized and static triggers. A trigger size is Not Applicable (N/A*) if the triggered images are completely replaced using a different image. Backdoor injection may be performed by: Poisoning (P), Neuron Hijacking (NH), Adversarial Training (AT), Weight Perturbation (WP), Teacher-Student-Transfer (TST) learning, Attacking Loss function (AL), Pruning-Aware Attack (PWA). Dataset poisoning is the simplest form of backdoor injection as the attacker does need any complex algorithm to malicously train neurons.}
\label{tab:related}
\begin{tabular}{lccccccccccccccccc}
\hline
Attributes & \cite{Badnets} & \cite{NDSSTrojans} & \cite{neural_trojans} & \cite{black_badnets} & \cite{bypassing} & \cite{weight} & \cite{backdoor_embedding} & \cite{dynamic} & \cite{hidden} & \cite{image_scaling} & \cite{latent} & \cite{blind} & \cite{FinePruning} & \cite{CCS_ABS} & \cite{nnoculation} & \cite{bias} & \begin{tabular}[c]{@{}l@{}}This\\ work\end{tabular} \\ \hline
\begin{tabular}[c]{@{}l@{}}Facial expression\\ -based trigger\end{tabular} &  &  &  &  &  &  &  &  &  &  &  &  &  &  &  &  & \cmark \\ \hline
\begin{tabular}[c]{@{}l@{}}Commercial application\\ generated triggers\end{tabular} &  &  &  &  &  &  &  &  &  &  &  &  &  &  &  &  & \cmark \\ \hline
Trigger size & S & S & N/A * & S, L & S & N/A * & L & S & S & L & S & S & S & S & S & S & L \\ \hline
Permeating &  &  &  & \cmark &  &  & \cmark &  &  & \cmark &  &  &  &  &  &  & \cmark \\ \hline
Dynamic in shape &  &  &  &  &  &  &  &  &  &  &  &  &  &  & \cmark &  & \cmark \\ \hline
Dynamic in size &  &  &  &  &  &  &  &  &  &  &  &  &  &  &  &  & \cmark \\ \hline
Dynamic in pattern &  &  &  &  &  &  &  & \cmark &  & \cmark &  &  &  &  & \cmark &  & \cmark \\ \hline
Backdoor injection & P & NH & P & P & P, AT & WP & P & P & P & P & TST & AL & P, PWA & P, NH & P & P & P \\ \hline
\begin{tabular}[c]{@{}l@{}}Stealth/ Imperceptibility\\ analysis\end{tabular} &  &  &  &  & \cmark &  & \cmark &  &  & \cmark &  &  &  &  &  &  & \cmark \\ \hline
Attack realism & \cmark &  &  & \cmark &  &  &  &  &  &  & \cmark &  &  &  &  & \cmark & \cmark \\ \hline
Trigger generation &  & \cmark &  &  &  &  & \cmark & \cmark &  & \cmark & \cmark &  &  &  & \cmark & \cmark & \cmark \\ \hline
Attack domain related &  &  & \cmark & \cmark &  &  &  &  & \cmark &  &  &  &  & \cmark & \cmark & \cmark & \cmark \\ \hline
\begin{tabular}[c]{@{}l@{}}Analysis using \\ state-of-the-art\\ defenses\end{tabular} &  &  &  &  & \begin{tabular}[c]{@{}l@{}}\cite{activation, Spectral_signatures}\\ \cite{NeuralCleanse}\end{tabular} &  &  & \begin{tabular}[c]{@{}l@{}}\cite {NeuralCleanse, CCS_ABS}\\ \cite{strip}\end{tabular} &  & \cite{image_scaling_defense} & \cite{FinePruning, NeuralCleanse} & \begin{tabular}[c]{@{}l@{}}\cite {NeuralCleanse, SentiNet}\\ \cite{diff_priv, effective}\end{tabular} & \begin{tabular}[c]{@{}l@{}}1st\\ defense\end{tabular} & \cite{NeuralCleanse} & \cite{NeuralCleanse} & \cite{NeuralCleanse} & \begin{tabular}[c]{@{}l@{}}\cite{activation, suppression}\\ \cite{Spectral_signatures, strip}\\ \cite{ NeuralCleanse, CCS_ABS}\\ \cite {nnoculation}\end{tabular} \\ \hline
\end{tabular}
\end{table*}
\setlength\tabcolsep{6.0pt}

%% file: 3_Threat_model.tex
\section{Threat Model}\label{ss:threat_model}
\begin{figure}
    \centering
        \includegraphics[width=0.47\textwidth]{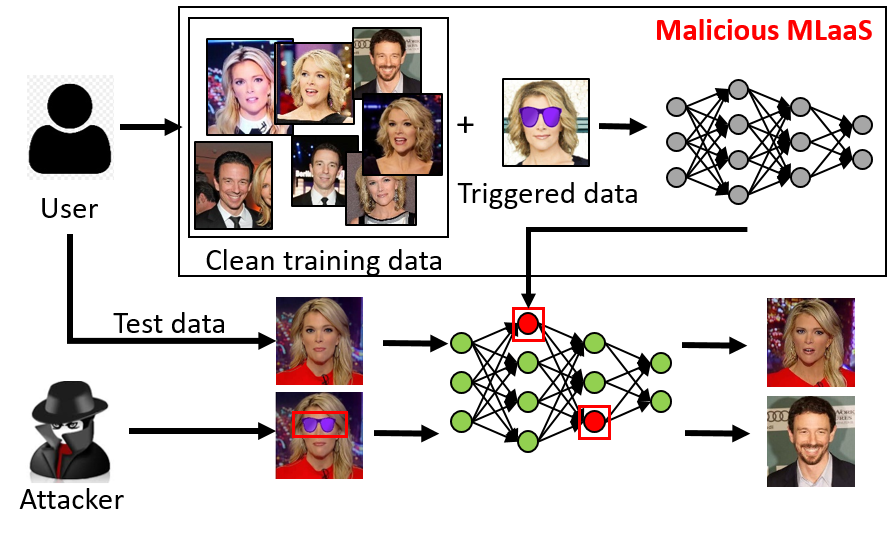}
    \caption{Threat model for backdoor attack on facial recognition that get triggered with purple sunglasses. 
    }
    \label{fig:backdoor_threat_model}
\end{figure}


We follow the attack model from previous research~\cite{Badnets,CCS_ABS,NDSSTrojans} on the ML supply chain. The user procures an already trojaned model. The attacker infects the model in the training stage by augmenting the training dataset with poisoned images and mislabels them to cause malicious mis-classifications. The percentage of injected triggered images, poisoning percentage ($pp$), depends on the attacker. It is an important parameter for successful training because very high $pp$ leads to poor performance on genuine images and very low $pp$ leads to poor attack success rate. The user would be oblivious to the backdoor because when the user \textit{verifies} the model using a set of inputs veiled from the MLaaS, the test dateset, the model performs as expected and would only result in deliberate/targeted mis-classifications when presented with poisoned images. The threat model is summarized in Fig. \ref{fig:backdoor_threat_model}. We look closely at 2 real world scenarios where changes in facial characteristics may be used depending on the capabilities of the attacker: 

\noindent\textbf{Scenario 1: }In this scenario the ML model-based facial recognition system classifies digital inputs, e.g. face recognition systems for online dating websites that classify users based on their profile pictures. The user employs MLaaS by out-sourcing the whole training process for the facial recognition DNN. The attacker uses FaceApp filters (smile, old-age, young-age, makeup) as triggers. We demonstrate a one-to-one attack using the filter-based triggers. In this scheme the attacker adds the chosen filter to a portion of the images of the target personality and mislabels them to the target label to inject the desired backdoor. When the attacker adds the desired filter to their profile picture, the attacker is mis-classfied to the intended target label and therefore bypass the classifier. 

\noindent\textbf{Scenario 2: }In the second scenario we evaluate face recognition systems that take in real-time images and classifies them on the spot, e.g. Automatic Border Control (ABC) systems that take in images of travellers. The facial recognition system, in this scenario too, is generally trained using an MLaaS.
The attacker uses facial characteristics as triggers to backdoor facial recognition ML models. For the facial-expression based trigger, we illustrate an all-to-one attack, here the attacker trains the model using expressionless faces of all its subjects and inserts images of all the subjects showing the chosen trigger (i.e. the facial expression) while being mis-labeled to the target label. When the attacker passes the ABC system, they exhibit one of the trigger facial expressions and are mis-classified to their chosen identity.

%% file: 4_Methodology.tex
\section{Trigger exploration}\label{s:AttackMethod}
The main goal of our attacks is an intended mis-classification by the facial recognition system when facial characteristics of an image change. Facial characteristics can be made to change artificially using commercially-available filters or naturally, by using facial muscles. Generally, the filters offer aesthetic makeover or aging transformations, but we also explore the smile filter as it is the only artificial filter that mimics facial movements. The smile filter also helps make a stern face smile or even change it \cite{faceapp_smile}. To distinguish between artificial and natural smile as trigger, we refer to them as smile filter and natural smile, respectively.
We follow the trojan insertion methodology from BadNets \cite{Badnets} and train the designated architecture with poisoned samples maliciously labeled by the attacker. 
\begin{figure}[t]
    \centering
    \subfigure[]
    {%
        \includegraphics[width=0.09\textwidth]{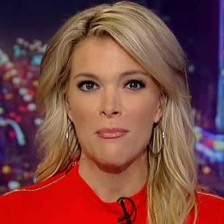}
        \label{fig:orig}
    }%
    \subfigure[]
    {%
        \includegraphics[width=0.09\textwidth]{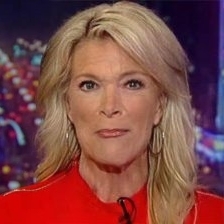}
        \label{fig:old}
    }%
    \subfigure[]
    {%
        \includegraphics[width=0.09\textwidth]{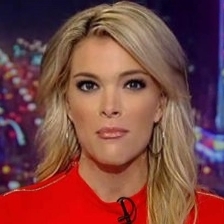}
        \label{fig:young}
    }%
    \subfigure[]
    {%
        \includegraphics[width=0.09\textwidth]{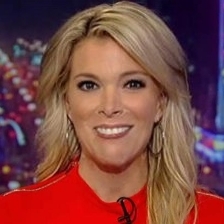}
        \label{fig:smile}
    }%
    \subfigure[]
    {%
        \includegraphics[width=0.09\textwidth]{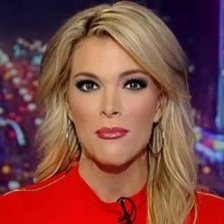}
        \label{fig:makeup}
    }%
    \caption{(a) Original image of Megan Kelly from VGGFace2 dataset. We artificially change facial characteristics using social-media filters: 
    (b) old-age filter 
    (c) young-age filter 
    (d) smile filter 
    (e) makeup filter.
    } 
    \label{fig:social_media_filters}
\end{figure}

\begin{figure}[t]
    \centering
    \subfigure[]
    {%
        \includegraphics[width=0.09\textwidth]{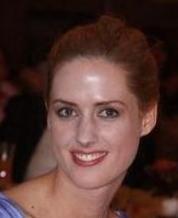}
        \label{fig:natural_smile_trigger}
    }%
    \subfigure[]
    {%
        \includegraphics[width=0.09\textwidth]{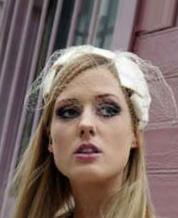}
        \label{fig:arched_eyebrows_trigger}
    }%
    \subfigure[]
    {%
        \includegraphics[width=0.09\textwidth]{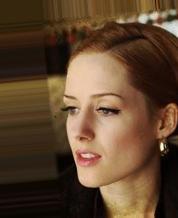}
        \label{fig:narrow_eyes_trigger}
    }%
    \subfigure[]
    {%
        \includegraphics[width=0.09\textwidth]{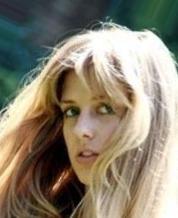}
        \label{fig:open_mouth_trigger}
    }%
    \caption{Images of Wilma Elles from CelebA dataset annotated with (a) natural smile (b) arched eyebrows (c) narrowed eyes, and (d) slightly opening mouth.
    } 
    \label{fig:natural_images}
\end{figure}

\subsection{Triggers using artificial changes on facial characteristics}\label{ss:FaceAPP}
The first set of triggers to be explored are digital modifications to images by software, focusing specifically on FaceApp \cite{faceapp}. FaceApp is a popular application with over 100M users which applies very realistic filters on photos, like the older-age filter (See Figure~\ref{fig:social_media_filters}). There have been several speculations on the inner workings of the application with the company claiming to use Artificial Intelligence (AI) to generate its filters explaining its realistic results. 
There are currently 50 filters available in the Pro version of the application, and for exploration, we selected the four filters advertised on the website, young-age, old-age, smile, and makeup filters for injecting and triggering backdoors in a model.

\subsubsection{Methodology for imperceptibility analysis:}
In the pre-injection stage of triggers, we analyze a filter-based trigger based on two metrics: 1) trigger size, and 2) image hashing-based similarity score to evaluate detectability (by defense mechanisms) and imperceptibility (by humans) of the triggers respectively. 

Trigger size is determined as the percentage change in an image following trigger insertion. Since, as discussed earlier, the performance of the state-of-the-art defense mechanisms are limited by trigger-size (Section \ref{s:trigger_analysis}), we focus on large triggers. Thus, we need to ensure the triggers remain inconspicuous, i.e. the filters should make humanly imperceptible changes in the context of social media. Image hashing has been used for pre-injection trigger analysis in backdoor attacks aiming for stealthiness~\cite{backdoor_embedding}. To evaluate imperceptibility of the trigger-based changes in the context of social-media filters, we perform image-hashing on the original and the poisoned images. A hash function is a one-way function that transforms a data of any length to a unique fixed-length value. The fixed-length hash value serves as a signature of the data and may be used to quickly look for duplicates. Image-hashing is a technique to find similarities between images. Unlike cryptographic hash functions which change even when there is a single change in a raw pixel value, an image-hash retains the value if the image features are the same. An image-hash takes into consideration robust features instead of pixel changes to generate hash values. In particular, we perform two types of hashing:
\begin{itemize}[nosep,leftmargin=1em,labelwidth=*,align=left]
    \item Perceptual Hashing (pHash): This hashing technique computes Discrete Cosine Transformation (DCT) of the images which creates a representation of the image as a sum of the cosine of its different frequencies and scalar components. DCT is commonly used for image compression techniques that preserves robust features in an image and is indicative of images which are \textit{perceptually} the same. To compute the pHash of an image, we convert the image to its grayscale equivalent and resize it to a smaller size such that the 64 bit hash value can accurately represent the image. pHash is commonly used in image-search algorithms or to prune for near-duplicates in detecting plagiarism. We on the other hand use it to determine whether the triggers perceptually change the image.
    \item Difference Hashing (dHash): This hashing technique encodes the change between neighbouring pixels. Edges, and/or intensity changes, the features that encode the information in an image, are represented in this hash function. Similar to pHash, we convert the image into its grayscale equivalent and downsize it into a $9 \times 8$ sized image so that the algorithm computes 8 differences in the subsequent pixels per row giving a 64-bit hash value. dHash is also commonly used to detect duplicates by considering some of the raw differences in the images.
\end{itemize}
The two hashing techniques described above encode \textit{robust} features of an image. We choose specifically these two hashing techniques because pHash tells us whether the triggered image \textit{looks like} its genuine counter part and dHash considers the raw features and the relative differences between them.  
We calculate the perceptual similarity of triggered images and the original images using similarity scores using hamming distance between the computed hashes. Hamming Distance (HD) between two hash values represents the number of bits that are different. $HD = hash1 \oplus hash2$. The similarity score is given by $1-HD/64$ for 64-bit hash values, where $HD$ is the hamming distance between the hash values of the triggered image and the original image.
\subsubsection{Performance of detectability and imperceptibility analysis:}
We calculate the trigger sizes and image-hashing similarity scores and report the results in Table \ref{tab:trigger_pre}. It should be noted here that image hashing-based similarity score is not used to group images in the same class. Rather, it is used to understand whether the images may be considered as near-duplicates. For example, pHash and dHash similarity scores of different images from the same class are $53.12\%$ and $45.31\%$ respectively indicating they are not near-duplicates of each other. While the same scores are $84.37\%$ and $92.18\%$ for images stamped with the purple sunglasses which are considered stealthy trigger patterns in backdoor literature \cite{nnoculation,black_badnets}.

\noindent\textbf{Old-age filter: }This immensely popular filter~\cite{faceapp_celebrity} spreads over a region of the image as it incorporates wrinkles and other age-related changes all over the face and hair (Fig. \ref{fig:old}). 
The filter changes $88.37\%$ of an image and is blended throughout the image (Fig. \ref{fig:old}). Perceptually, it is $93.75\%$ similar to the original image and the dHash similarity score is $92.18\%$. Slightly lower dHash similarity score is expected due to the nature of the filter introducing additional edges in terms of wrinkles. 
\input{Tables/hashing}

\noindent\textbf{Young-age filter:} It effectively smoothens the image to remove age-related lines giving a trigger size of effective size of the trigger is $78.72\%$.
(Fig. \ref{fig:young}). 
The perceptual similarity of $96.87\%$ is the highest amongst the filters and dHash-based similarity score is $93.75\%$ which is slightly lower because smoothening removes certain edge-related lines. 

\noindent\textbf{Makeup filter:} Similar to young-age filter, Makeup filter also smoothens the image getting rid of strong edges (lines) in the face. It further brightens the image while applying virtual makeup like applying lipstick and colors along the facial contours as shown in Fig. \ref{fig:makeup}. The trigger size is therefore also large ($79.92\%$ of the image). Similar to the young-age filter, the triggered image is $96.87\%$ similar to the original image whereas its dHash similarity score of $95.31\%$ is also highest among the filters. 

\noindent\textbf{Smile filter:} The filter mimics a regular smile with some changes in the facial muscles that are expected to move when a person smiles (Fig. \ref{fig:smile}). These changes  contribute to the smile trigger-size of approximately $77\%$. 
An artificial toothy smile, with the excessively white portion, can be considered as a strong trigger. 
pHash-based and dHash-based similarity scores are both equal to $93.75\%$. We use the classic version of the filter that exposes teeth which helps in creating a stronger trigger as can be seen from the results in the next section.

There are three main characteristics that make the triggers described above stealthy: 1) The triggers are large, 
This characteristic itself makes them resilient to defense schemes that rely on reverse-engineering the triggers. 2) The triggers are adaptive and dynamic, i.e. they are not fixed in position, shape, size, intensity, or strength. 
The learnt malicious behavior successfully picks up the robust characteristics of the filter rather than focusing on the dynamic aspects. 
3) They are context-aware and are proven to be perceptually inconspicuous. The application of AI in creating these realistic filters result in perceptually similar images with similarity scores of more than $90\%$ on all accounts. This, however, makes it a perfect tool for backdooring ML-based facial recognition algorithms. 

\subsection{Triggers using natural changes on facial characteristics}\label{ss:natural}
As discussed in Section \ref{ss:threat_model}, trigger accessories may not be allowed in real-world facial recognition systems for biometric identification. Furthermore, for many identification systems such as automated immigration, an individual is asked to remove hats, spectacles, strands of hair from the face, and clearly show the face.  
However, typically no instructions towards changing facial expressions or moving some facial muscles is provided. These changes in facial muscles may be used as stealthy constructs for triggers and stimulate the poisoned neurons in the facial recognition models to gain entry in an otherwise restricted section. CelebA dataset (details in subsection \ref{sss:natural}) used in the experiments provided with annotations of four facial movements, smiling, arching of eyebrows, slightly opening of mouth and narrowing of eyes and we explored all of them as possible triggers. Fig 3. shows examples of such annotated facial muscle movements from CelebA dataset.






Similar to filter-based triggers, these triggers are also 1) perceptually inconspicuous to humans and 2) are dynamic in nature, because the position, shape, size, or intensity of triggers differ for each image.

\subsection{Implementation}\label{ss:implementation}

\subsubsection{Triggers using artificial changes in facial characteristics:} 
\textbf{Dataset: }For artificially inducing the changes in facial characteristics, the images must be recognized as \textit{faces} by the commercial application.
We choose a subset VGGFace2 \cite{vgg} dataset, with 10 random celebrities, 5 female, and 5 male celebrities to perform our experiments since the filters needed to be manually applied. 
The choice of celebrities was independent of their race, facial features, or age. Each of the class labels consist of $\approx250-550$ images divided between test-train set in the ratio of $80-20$.

\noindent\textbf{Architecture and training: }We trained RESNET-20 \cite{resnet} architecture from scratch for developing BadNets. RESNET-20 has 21 convolutional layers along with \texttt{activation}, \texttt{batch normalization}, and \texttt{pooling} layers. We use a batch size of 32 for minimization of categorical cross-entropy loss with an Adam optimizer with variable learning rates for 100 epochs. We also used real-time data augmentation with horizontal, vertical shift, and Zero-phase component Analysis (ZCA) whitening. Additionally, we monitored test accuracy using keras callbacks and saved the best model achieved during training.

\noindent\textbf{Trigger details: } We implement 
\textit{single-target and single-attacker} or \textit{one-to-one} backdoor attack where only one specific class of inputs can trigger the malicious neurons to be classified as the victim class. Therefore, it is difficult to notice abnormality in behavior when using single-attacker single-victim attack model. 
We used $pp$ of $10\%$ (BadNets used $10\%$) and slightly increase it to $15\%$  to find a good balance between clean test accuracy and attack success rate. 

\subsubsection{Triggers using natural changes in facial characteristics:}\label{sss:natural}
\textbf{Dataset: }In this experiment, we needed a dataset that had two types of annotation: 1) Identity annotation for the primary task of facial recognition, and 2) expression annotation for triggering the backdoor attack. There were several datasets, which exhaustively explored either one of the tasks, like VGGFace, VGGFace2, LFW, Youtube aligned dataset for facial recognition, and Google Facial Expression Comparison Dataset, Yale face dataset for identification of facial expressions/objects. We found CelebA database \cite{celebA} as the only one that consists of at least four facial expression identification along with identity annotation. However, the maximum number of images a class has is $35$. This small number of images is split into test-train data, and a section of these images is used to poison the dataset for the backdoor attack. Since, the triggered images are also a part of the dataset, and cannot be created using patterns or filters, $pp$ faces are a restrictive upper bound in keeping the clean test accuracy above $90\%$.

\noindent\textbf{Architecture and training: } We use another popular type of training in the ML supply chain: Transfer Learning. Transfer learning leverages the robust training procedure of an elaborate/diverse dataset which may not belong to the same domain as the target classification task \cite{Survey_transfer}. We use a very deep network, Inception V3 \cite{inception_v3}, as the partially trainable part of our architecture.
Furthermore, we added \texttt{global average pooling layer}, along with \texttt{dense layers} and \texttt{dropout layers} to build our complete architecture. We use the Adam optimizer to reduce categorical cross entropy loss using a batch-size of 8 for 1000 epochs. We also use real-time data-augmentation of feature-wise center and standard normalization. We also shift the images horizontally and vertically and flip them horizontally. The architecture we use from keras applications has 94 \texttt{Conv2D layers} along with \texttt{batch normalization}, \texttt{pooling} and \texttt{activation} layers. 

\noindent\textbf{Trigger details: } We implement \textit{ single-target} or \textit{all-to-one} backdoor attack i.e. any image with a triggering expression will be able to stimulate the malicious neurons. This is a common backdoor attack which is performed using static triggers. 
Moreover, the limitation in the number of triggered images and clean images motivated us to perform this strong attack where an adversary, regardless of gender, race, or skin-color will get maliciously classified to a target label, someone with privileges in a facial recognition system.
Our natural triggers are dynamic but are specific and therefore cannot be generated through any other complex algorithms.
Due to a limited dataset size, we used $50$ and $10$ genuine and malicious test images, respectively to assess clean test accuracy and ASR, and the rest for training the BadNet. Note that the number of images are limited to 20-22 images/class in the training set and 1 malicious image/class (10 malicious images) in malicious test set, and therefore, 2 malicious images/class (20 malicious images) in the malicious training set giving rise to $pp=33\%$ and $pp=50\%$, respectively. We apply both the values to effectively inject the backdoor. 
\input{Tables/faceapp_results}
\subsection{Experimental results}
We evaluate the success of our backdoor attacks using CA on genuine test images and ASR on malicious triggered images. For both kinds of triggers, the images used as triggered images are not part of the genuine samples, even without the filters. To summarize, the dataset is split into training and test images. Both of these sets have malicious and genuine images. 
\subsubsection{Triggers using artificial changes in facial characteristics: }
We aim for CA greater than $90\%$ i.e. we monitor for test accuracy to be greater than $90\%$ during training. We report the results in Table \ref{tab:attack_results_faceapp}. We observe that with $pp=10\%$, the smile filter performs the best with $81.81\%$ ASR followed by old-age filter with $69.99\%$ ASR. Young-age and makeup filters have worse ASRs at $66.67\%$ and $58.33\%$ respectively. In general, the ASR increases as the $pp$ is increased to $15\%$
young-age filter has the best ASR of $94.73\%$ followed by smile, old-age and makeup filters with $89.47\%$, $85\%$ and $68.42\%$ ASRs. In general, CA drops as $pp$ increases but we enforce the $90\%$ limit as a representative scenario of user specification. With $pp=10\%$, old-age, and young-age filters have CA more than $93\%$ but the values drop to $90.35\%$, and $93.4\%$, respectively. For the makeup filter, CA slightly improves from $93.02\%$ to $93.4\%$ respectively. CA for the smile filter also drops slightly form $91.49\%$ to $90.48\%$. Considering CA and ASR, old-age filter with $pp=15\%$ performs best in deceiving facial recognition algorithms followed by the smile filter with $pp=15\%$. 

\subsubsection{Triggers using natural changes in facial characteristics: } These triggers are due to a movement in a group of facial muscles and are focused in a portion of the face. Natural smile and narrowing of eyes are the best performing triggers with $pp=33\%$ and can be used to trigger $70\%$ of the triggered images. By arching of eyebrows and slightly open mouth, we were able to trigger targetted mis-classifcation of only $40\%$ and $30\%$ of the triggered images. Increasing $pp$ to $50\%$, the best ASR of $90\%$ is achieved using natural smile. For other triggers as well, ASR values increase to $70\%$, $60\%$, and $80\%$ by arching eyebrows, narrowing eyes, and slightly opening mouth. Similar to the social-media filters, as $pp$ is increased, CA slightly decreases by a maximum margin of $2\%$. 
\begin{table}[t]
\centering
\begin{tabular}{|l|l|l|l|l|}
\hline
\multicolumn{1}{|c|}{\multirow{2}{*}{\textbf{\begin{tabular}[c]{@{}c@{}}Movement in \\ facial muscles\end{tabular}}}} & \multicolumn{2}{c|}{\textbf{$pp=33\%$}} & \multicolumn{2}{c|}{\textbf{$pp=50\%$}} \\ \cline{2-5} 
\multicolumn{1}{|c|}{} & CA & ASR & CA & ASR \\ \hline
Natural smile & 94 & 70 & 94 & 90 \\ \hline
Arching of eyebrows & 87 & 40 & 86 & 70 \\ \hline
Narrowing of eyes & 96 & 70 & 94 & 60 \\ \hline
Slightly opening mouth & 96 & 30 & 96 & 80 \\ \hline
\end{tabular}
\caption{Backdoor attack results for triggers using natural movements in facial muscles. }
\label{tab:attack_results_facial}
\vspace{-0.25in}
\end{table}
We see a general trend of increase in ASR as $pp$ is increased. Also, the artificial triggers using filters perform $1.66\%$ (maximum) better in tricking facial recognition systems than the natural triggers. Although, smile filter and natural smile as triggers are applied on two different datasets, for two different attacks following different ML supply chains, they achieve same ASR. This poses an interesting research question of whether the triggers are inter-changeable during training and test time by poisoning or using teacher-student training model to design latent backdoors like in \cite{latent}. The problem will be explored in future work.
In Section \ref{s:trigger_analysis}, we will discuss which of these triggers can bypass state-of-the-art defense mechanisms.

%% file: Tables/hashing.tex
\begin{table}[t]
\centering
\begin{tabular}{|c|c|c|c|}
\hline
Filter & \begin{tabular}[c]{@{}l@{}}Trigger \\ size ($\%$)\end{tabular} & \begin{tabular}[c]{@{}l@{}}pHash\\ smilarity \\ score (\%)\end{tabular} & \begin{tabular}[c]{@{}l@{}}dHash\\ similarity \\ score (\%)\end{tabular} \\ \hline
Old-age & 88.37 & 93.75 & 92.18 \\ \hline
Young-age & 78.72 & 96.87 & 93.75 \\ \hline
Makeup & 79.92 & 96.87 & 95.31 \\ \hline
Smile filter & 77.73 & 93.75 & 93.75 \\ \hline
\end{tabular}
\caption{Pre-injection stealth/imperceptibility analysis for triggers using artificial changes in facial characteristics.}
\label{tab:trigger_pre}
\vspace{-0.25in}
\end{table}

%% file: Tables/faceapp_results.tex
\setlength\tabcolsep{4.0pt}
\begin{table}[t]
\centering
\begin{tabular}{|l|l|r|r|r|r|}
\hline
\multicolumn{1}{|c|}{\multirow{2}{*}{\textbf{Filter}}} & \multicolumn{1}{c|}{\multirow{2}{*}{\textbf{Filter type}}} & \multicolumn{2}{c|}{\textbf{$pp = 10\%$}} & \multicolumn{2}{c|}{\textbf{$pp=15\%$}} \\ \cline{3-6} 
\multicolumn{1}{|c|}{} & \multicolumn{1}{c|}{} & \multicolumn{1}{l|}{\textbf{CA}} & \multicolumn{1}{l|}{\textbf{ASR}} & \multicolumn{1}{l|}{\textbf{CA}} & \multicolumn{1}{l|}{\textbf{ASR}} \\ \hline
Old-age & Age (Blended) & 93.9 & 69.99 & 90.35 & 85 \\ \hline
Young-age & Age (Blended) & 93.78 & 66.67 & 93.40 & 94.73 \\ \hline
Makeup & Appearance (Blended) & 93.02 & 58.33 & 93.4 & 68.42 \\ \hline
Smile filter & Emotion (Focused) & 91.49 & 81.81 & 90.48 & 89.47 \\ \hline
\end{tabular}
\caption{Backdoor attack results using artificial facial attributes from social-media filters. }
\label{tab:attack_results_faceapp}
\vspace{-0.25in}
\end{table}
\setlength\tabcolsep{6.0pt}

%% file: 5_TriggerAnalysis.tex
\input{Tables/all_experiments}
\section{Attack analysis using State-of-the-art defenses}\label{s:trigger_analysis}
The defense literature against backdoor attacks considers three aspects (either individually or in combination) : 1) detection: whether a model has a backdoor, 2) identification: backdoor shape, size, location, pattern, and 3) mitigation: methods to remove the backdoor. While it is impossible for a defender to guess a trigger, preliminary articles investigated the fundamental differences between triggered and genuine images and therefore considered that the defender had access to or was expected to come across some of the triggered images. Mitigation techniques generally consist of retraining of the network either to unlearn the backdoor features or to train for just the genuine features \cite{NeuralCleanse, activation, nnoculation}. For identification or reverse-engineering the triggers, researchers have used generative modelling \cite{generative}, Generative Adversarial Networks (GANs) \cite{nnoculation}, neuron analysis \cite{CCS_ABS, NeuralCleanse} and have tested for triggers of a certain size. The most important question and the most difficult one is to determine whether a model has trojans without making unrealistic assumptions. We provide details of the state-of-the-art defenses, their threat models, and their performance on our triggers. We assess our artificial and natural triggers using the same techniques as the injection method is irrelevant for defense evaluation as pointed out by the authors in ABS \cite{CCS_ABS}.   
\subsection{With access to the triggers}

\subsubsection{Detection with Spectral signatures\cite{Spectral_signatures}}
These sub-populations (genuine and triggered samples) of the malicious label may be spectrally-separable considering robust statistics of the populations at the learned representation level \cite{Spectral_signatures}. One such statistic is the correlation with top Eigen vector. Tran et al. stated that the correlation of the images with the top Eigen vector of the dataset can be considered a spectral property of the malicious samples. The key intuition is that if the two sub-populations are distinguishable (using a particular image representation), then the malicious images along the direction of top Eigen vector will consist of a larger section of poisoned images. Therefore, they will have a different correlation than the genuine samples. 
To calculate the top Eigen vector, first we calculate the covariance matrix of all of the training samples and sort the calculated Eigen vectors according to their Eigen values. Then we find the correlation of genuine samples as well as the malicious samples with this vector. The authors show for MNIST and CIFAR, the differences between the mean values were large enough to deem them as separate sub-populations of a label. Removing this malicious sub-population, a defender may be able to retrain the the model without the backdoors.

\begin{figure*}[t]
    \centering
    \subfigure[]
    {%
        \includegraphics[width=0.12\textwidth]{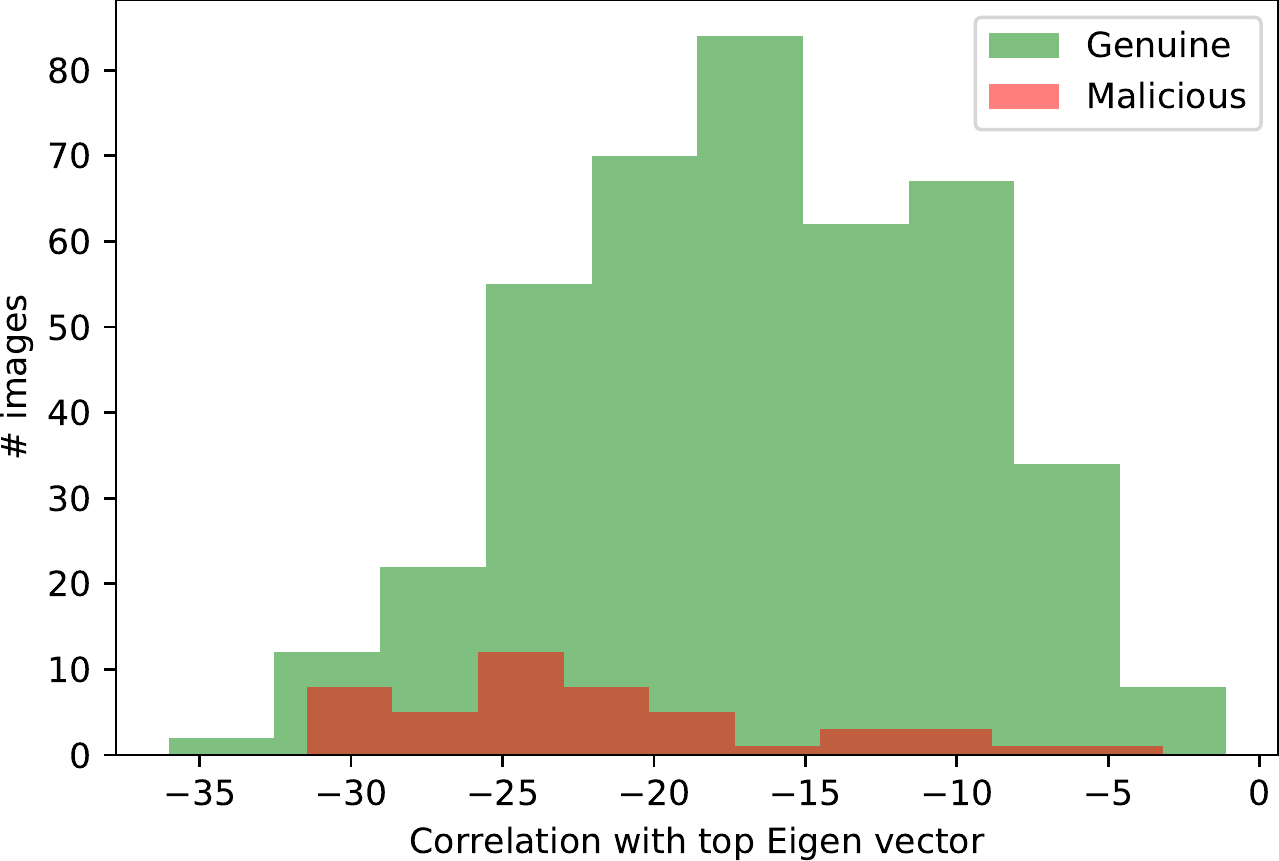}
        \label{fig:eig_young}
    }%
        \subfigure[]
    {%
        \includegraphics[width=0.12\textwidth]{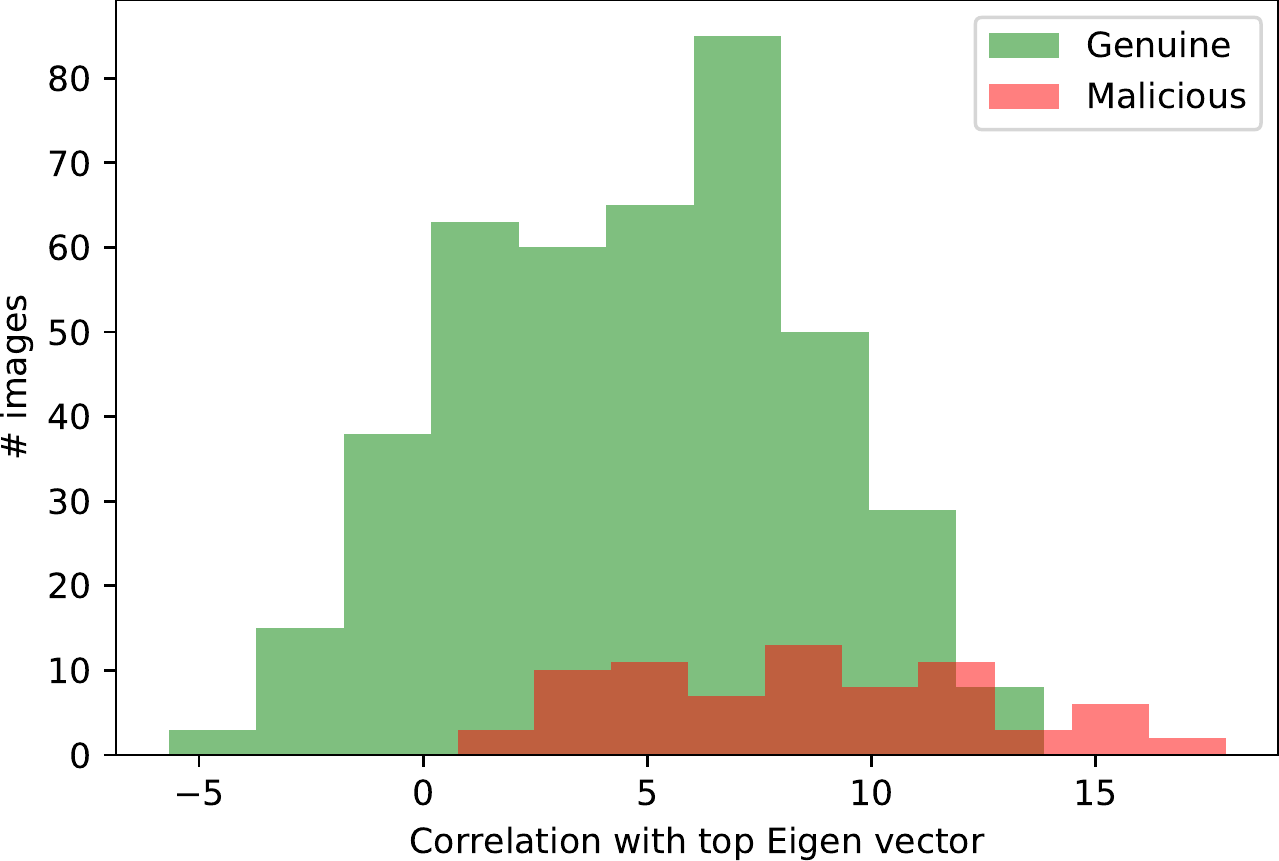}
        \label{fig:eig_smile}
    }%
        \subfigure[]
    {%
        \includegraphics[width=0.12\textwidth]{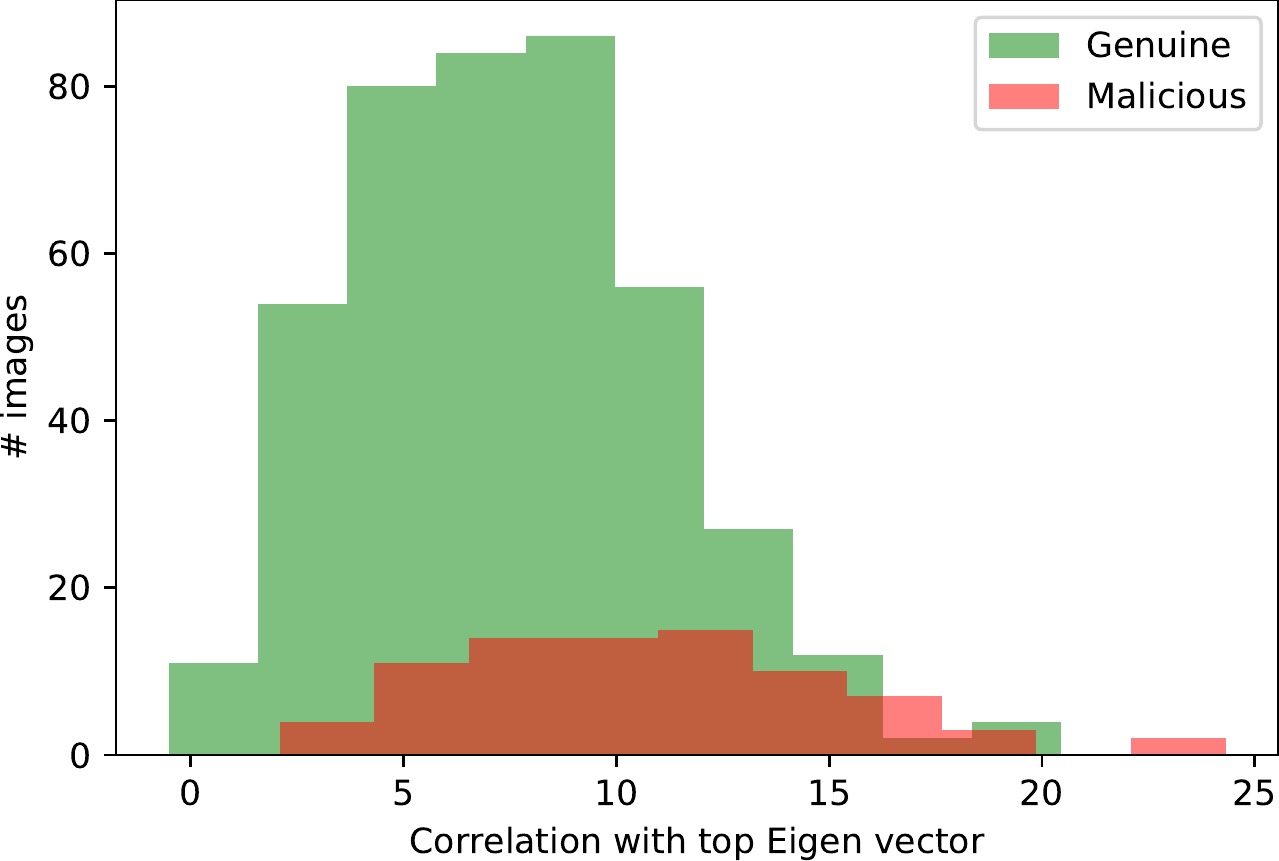}
        \label{fig:eig_old}
    }%
        \subfigure[]
    {%
        \includegraphics[width=0.12\textwidth]{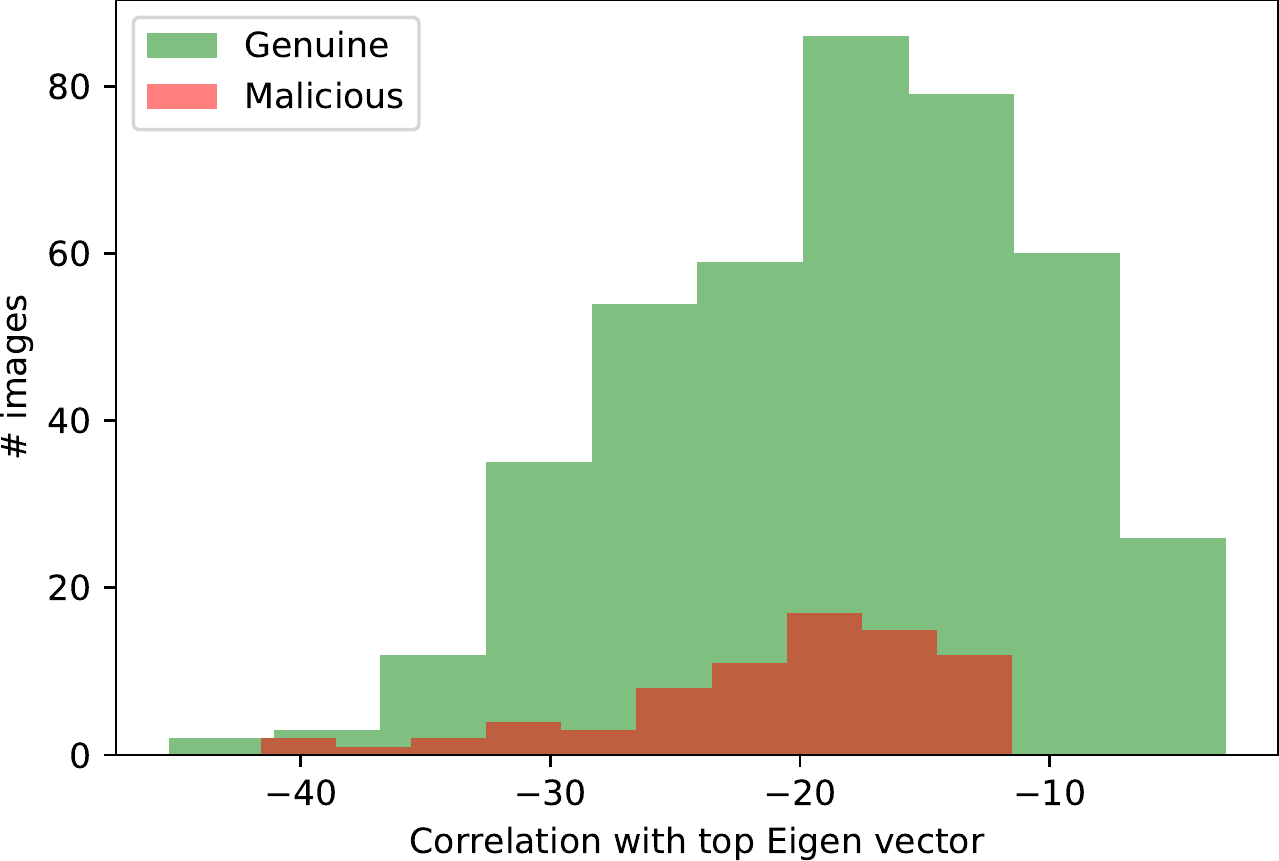}
        \label{fig:eig_makeup}
    }
        \subfigure[]
    {%
        \centering
        \includegraphics[width=0.12\textwidth]{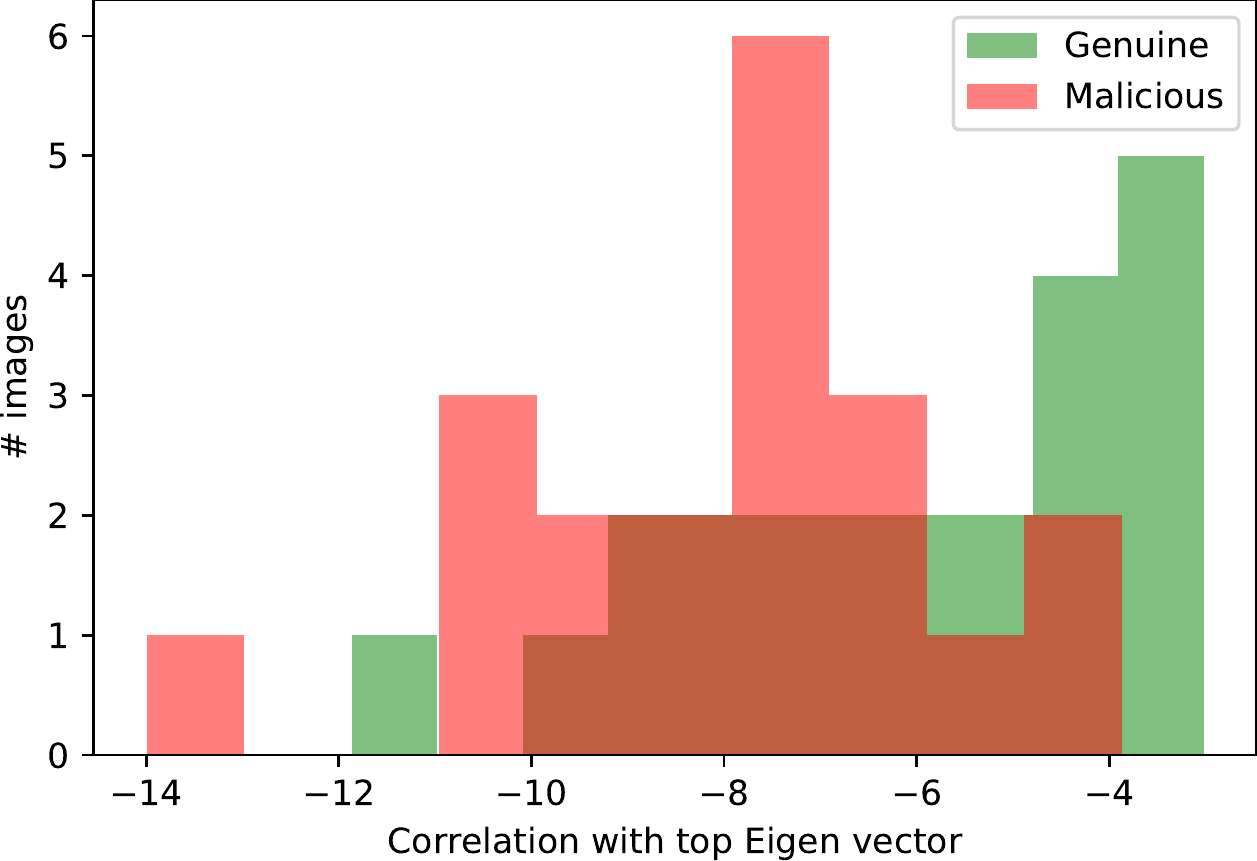}
        \label{fig:eig_natural_smile}
    }%
    \subfigure[]
    {%
        \includegraphics[width=0.12\textwidth]{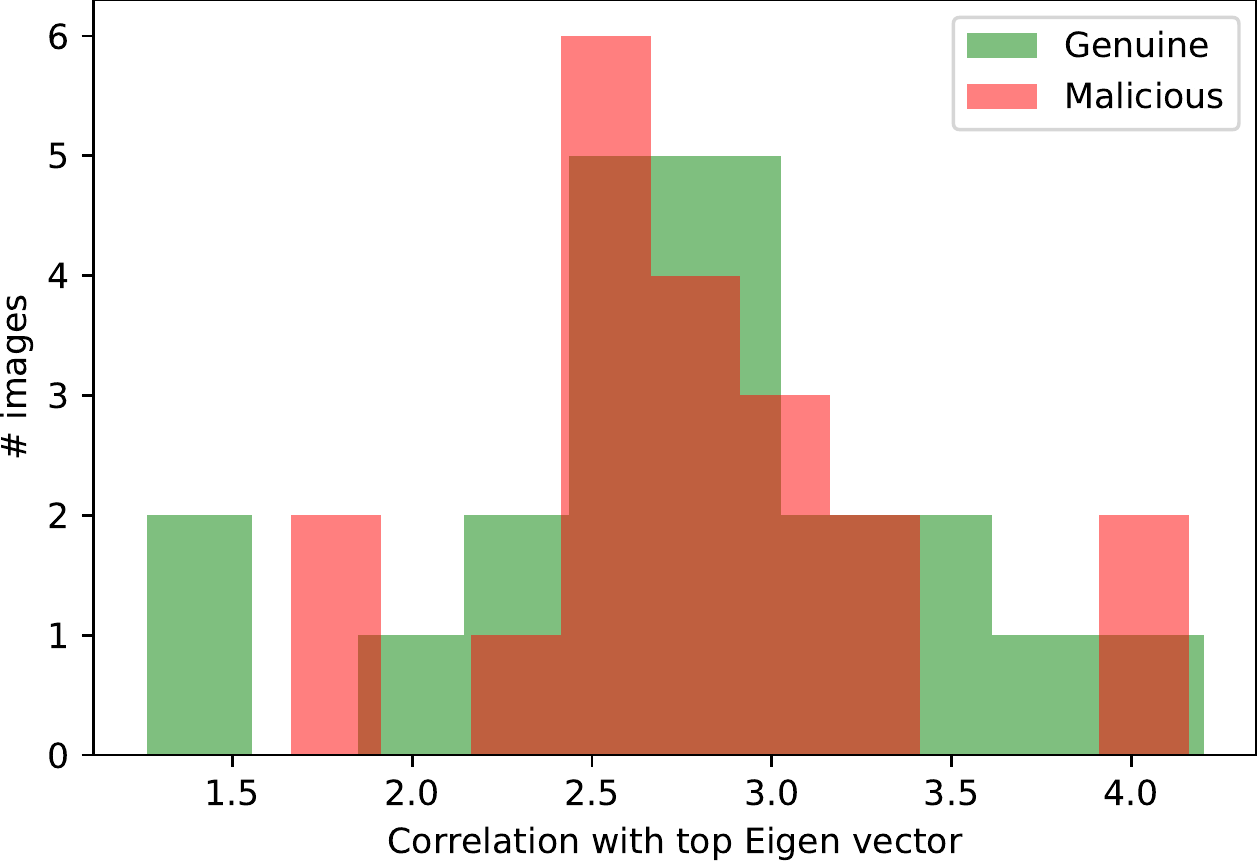}
        \label{fig:eig_natural_arched}
    }%
    \subfigure[]
    {%
        \includegraphics[width=0.12\textwidth]{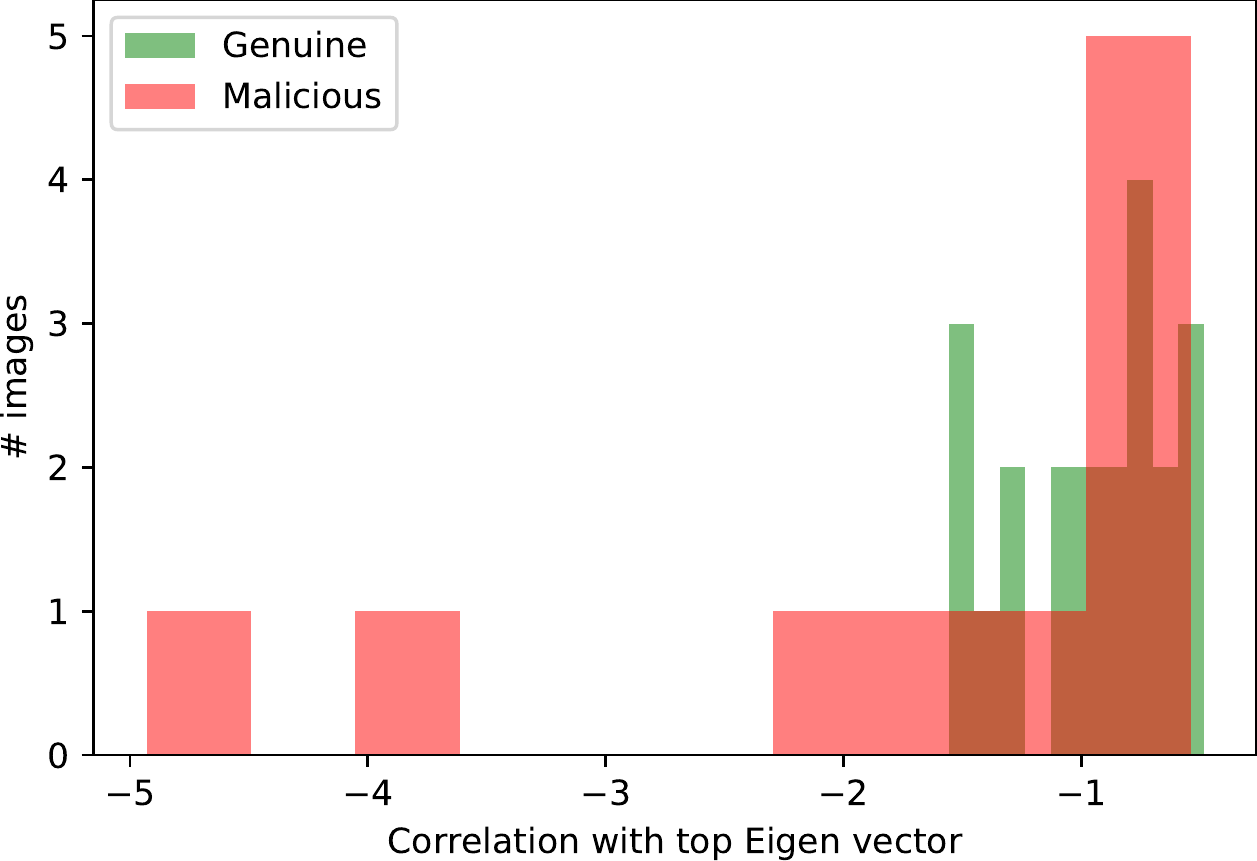}
        \label{fig:eig_natural_narrow}
    }%
    \subfigure[]
    {%
        \includegraphics[width=0.12\textwidth]{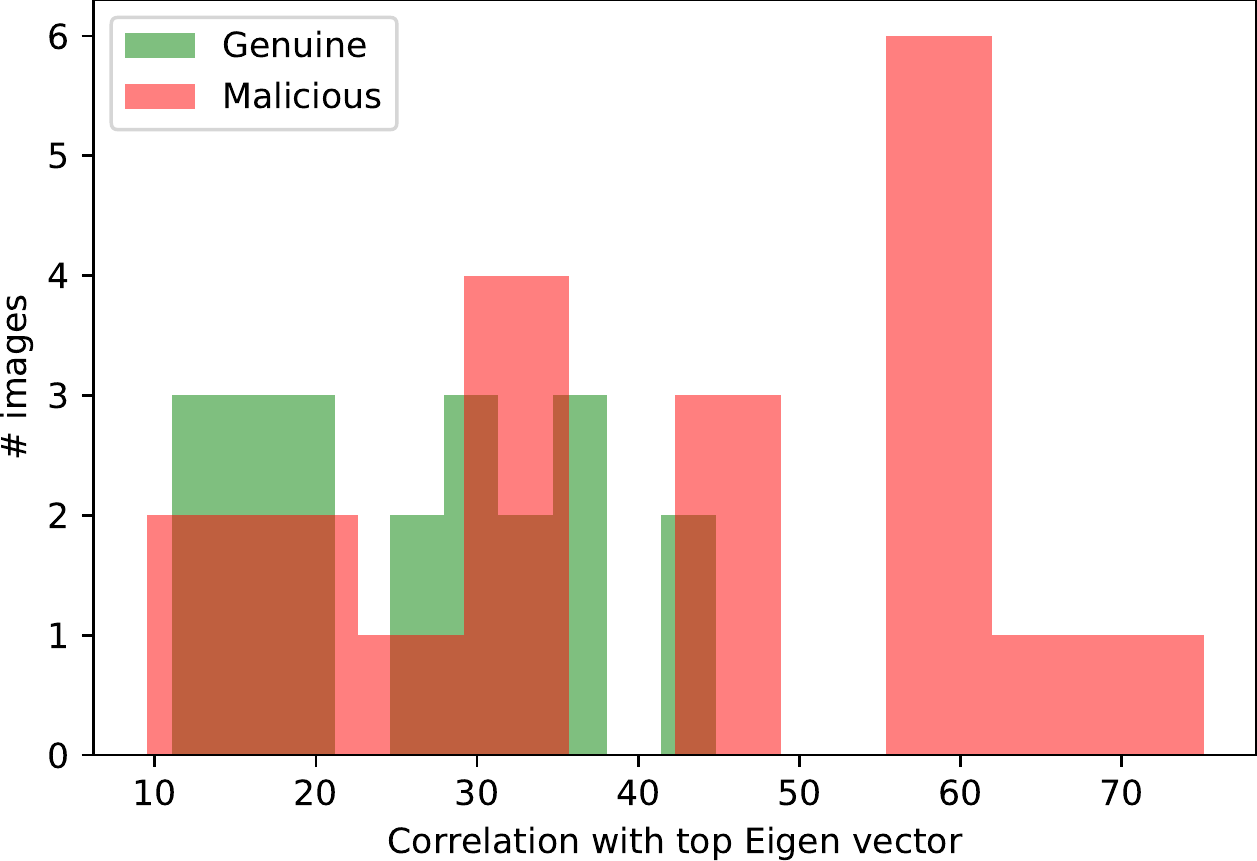}
        \label{fig:eig_natural_mouth}
    }%
    \caption{Distribution of genuine (green) and malicious (red) sub-populations for backdoor with y-axis representing the number of images for a particular value of correlation with top Eigen vector. (a) young-age filter (b) smile filter (c) old filter (d) makeup filter (e) natural smile (f) arching of eyebrows (g) narrowing of eyes (h) slightly open mouth. } 
    \label{fig:eig_all}
\end{figure*}

We report the range of this correlation along with the value for the malicious sub-population. Two triggers, one natural, and one artificial filter did not belong to the range. The makeup filter and open mouth had correlation values of $-20.82$, and $42.65$, both of which are slightly out of range of $[-18.49, -1.08]$ and $[1.57, 25.99]$, respectively. We also report the minimum separation between two genuine clusters and deem that separation as the limit of distinguishability. Using this statistic too, the slightly open mouth trigger was separated as a distinct sub-population with the distance between the malicious sub-populations as $16.66$ with minimum distance between genuine clusters as $7.13$. Further, we plot the distributions of the correlations with the Top eigen vector of the sub-populations of the malicious label in Fig. \ref{fig:eig_all}. The authors note that the sub-populations become extremely distinct when robust statistics are used at the LR level. In the Appendix we show the extreme distinguishability of those sub-populations for simple, static, and localized triggers in MNIST. Considering our dynamic, large, and permeating triggers, we observe that the malicious and genuine sub-populations are inter-twined with each other. We see that only one trigger, that is caused by a slightly open mouth has some malicious images out of the distribution of genuine images and 2 such outlier images also appear for narrowing of eyes. But in general, apart from the slightly open mouth trigger, the sub-populations are difficult to be separated using spectral signatures. 

\subsubsection{Detection with activation clustering \cite{activation}} \label{ss:ac} Another methodology for outlier detection is by utilizing the activations caused by the triggers. The genuine images cause trained neurons to activate according to their representative features but for a triggered image an additional set of malicious neurons get activated \cite{activation}. Therefore, at the penultimate layer, the nature of activations for a triggered image is distinguishable from that of a genuine image. Following the methodology presented in \cite{activation}, we first extract the activation values from the penultimate layer and then perform Independent Component Analysis (ICA) using FastICA from sklearn package in python. ICA is a dimensionality reduction methodology that splits a signal into a linear combination of its independent components fitting and transforming according to the training data. Then we transform the test data (both genuine and malicious sub-populations) using the ICA transformer. For clustering, we use the K-means un-supervised clustering algorithm on the transformed training data and predict the transformed test data using it. Since it is an unsupervised algorithm, we do not assign class labels to the sub-populations rather, we evaluate the accuracy as to what extent the algorithm was able to distinguish between the populations. Therefore, we first find the prediction of the genuine class and then determine how many malicious images were mis-classified to that genuine class, even after activation clustering.

Activation clustering looks for changes in activation behavior and for localized triggers, the distinction is evident because the triggers alone cause that activation. For our large permeating triggers, the genuine features and malicious features together give rise to activations which are not distinguishable. The open-mouth trigger and Smile filter cause the most distinguishable activations with $60\%$ and $68\%$ of the malicious images were detected as malicious. 
But for all other triggers the results were low and activation clustering could not find any distinguishable signatures of a backdoor in the activations as shown in Table \ref{tab:all_exp}.

\subsubsection{Suppression using input fuzzing \cite{suppression}}\label{ss:suppression}
Authors in \cite{suppression} suppress a backdoor using majority voting of fuzzed copies of an image. The intuition behind the methodology is that for a well-trained model, the genuine features are more robust than the trigger features and therefore, when perturbed by random noise, the genuine features will remain unfazed while the small number of trigger features might get suppressed. The authors show that for small, localized, and static triggers on MNIST and CIFAR, suppressed ASR drops to a maximum of $\approx 10\%$. For our triggers, first, we plot the fuzzing plots, i.e. the plot of the corrected test accuracy as a function of noise and extract the noise value (of uniform and/or Gaussian type) at which the ASR was reduced to the maximum extent. Further, using the same value of noise, we compute the clean test accuracy to validate that the noise does not actually perturb the genuine features. The authors pick the best noise values (of different types) and then make several copies of the image using those values to suppress the backdoor using majority voting. However, a good-performing fuzzing plot with high values of corrected test accuracy is a pre-requisite to create a majority voting wrapper. Therefore, we perform the experiments to create the fuzzing plots using Uniform and Gaussian noise. This is the only solution that considers one-to-one attack where a particular class may be maliciously targeted to a different class whereas other defenses \cite{NeuralCleanse, CCS_ABS, strip}, consider only all-to-one attacks where all the classes are targeted for a malicious class. Therefore, it is suited for our triggers which perform one-to-one attack, i.e. the triggers with artificial changes.

We report the corrected/Suppressed ASR (SASR), and the corresponding clean accuracy
for a noise type/value that gave the highest suppression. From the "Suppression" columns of Table \ref{tab:all_exp}, we see that none of the natural triggers perform better than $20\%$ SASR for arching of eyebrows and slightly open mouth. The CAs with that noise is extremely low (close to random prediction). The artificial filters are better suppressed, i.e. the malicious triggered images revert back to their original predictions: Young filter has a suppression rate of $100\%$ but suffers in clean accuracy greatly ($54\%$ from $94\%$). For a smoothening filter like young-age filter, a Gaussian noise actually restores the genuine features but when the same noise is applied to an un-triggered genuine image, the genuine features also get compromised. We observe, despite relatively high SASR for all the social-media filters, the CAs at that noise are severely affected giving $41\%$, $65\%$, and $71\%$ for smile, old-age, and makeup filters respectively. Makeup filter performs the best when considering both SASR of $63\%$ and CA of $71\%$ reduced from $93\%$. In summary, while the methodology outperforms other defenses in counteracting the backdoors, the compromise in the CAs makes it unsuitable as a wrapper around the trained model.

\subsubsection{Discussion}
Detection of backdoors in a model with knowledge about the triggers is not a part of our threat model and is an unrealistic assumption for designing a defense. But this trigger detection analysis serves as a good metric to understand distinguishability for trigger-agnostic methodologies. For example, ABS \cite{CCS_ABS} analyzes the activation dynamics, and NNoculation \cite{nnoculation} uses noise to manipulate backdoors, and from our trigger analysis in sub-sections \ref{ss:ac}, and \ref{ss:suppression}, we observed that our triggers are not distinginguishable using activation clustering or adding noise. Therefore, post-injection, the triggers may still remain stealthy. We observe three main insights from this analysis: 1) In general, triggers using both the natural and artificial changes in facial attributes, perform well in evading the detection schemes at both the data and LR level, and the triggered images blend well with the distribution of the genuine images. 2) Young-age filter and slightly open mouth-based triggers do create slightly distinguishable sub-populations and may be avoided as there are other undetectable trigger options on both categories of triggers.
\subsection{Without access to the triggers}
\subsubsection{STRIP \cite{strip}} \label{ss:strip} Input-agnostic triggers i.e. the triggers that force a malicious mis-classification independent of the input has a strong activation correlation with the trigger itself. Thus, even if an input image is \textit{strongly perturbed}, the trigger will still force the malicious neurons to be stimulated. Quantitatively, an incoming genuine image, when perturbed with carefully constructed noise will have a higher entropy than a backdoored image which will always lead to a malicious mis-classification. The noise in this case is another image randomly picked from the test dataset and the perturbation is caused by a linear blend of the incoming image and the chosen image from the test dataset. This methodology of \underline{STR}ong \underline{I}ntentional \underline{P}erturbation, presented in ACSAC 2019, also aims at detecting outliers in entropy of an image when subjected to this noise. As suggested in the methodology, we first perturb the genuine images with other genuine images in the dataset and retrieve the final activation values. Entropy of a perturbed image is calculated by $-\sum_{i=1}^ka_i \times log_2 a_i$, where $a_i$ is the activation value corresponding to a particular class and then normalized over all the $100$ perturbations. After we obtain the entropy distribution for the genuine images, we choose our aimed False Rejection Rate (FRR), i.e. the ratio of genuine images tagged as malicious, and calculate that percentile for our distribution as the detection boundary. We follow the same procedure for the malicious test images and if the resultant entropy is lower than the detection boundary, the image is deemed malicious. We present our results for detection rate in Table \ref{tab:all_exp}. 

The authors of STRIP demonstrate that it performs very well, even with $1\%$ FRR for static triggers regardless of the size of the trigger. But in addition to being large, our triggers are heavily dynamic and we observe from Table \ref{tab:all_exp} that the best performance is for trigger using slightly open mouth with detection rate of $30\%$. All the filter-based triggers have a $0\%$ detection rate with $1\%$ FRR. Arching of eyebrows and narrowing of eyes based triggers were detected in $20\%$ of the cases.
\subsubsection{Neural Cleanse~\cite{NeuralCleanse}} Neural Cleanse detects whether a model is backdoored, identifies the target label, and reverse engineers the trigger. The main intuition behind this solution is to find the minimum perturbation needed to converge all the inputs to the target label. Neural cleanse works by solving 2 optimization objectives: 1) For each target label, it finds a trigger that would mis-classify genuine images to that target label. 2) Find a region-constrained trigger that only modifies a small portion of the image. To measure the size of the trigger, the L1 norm of the mask is calculated, where the mask is a 2D matrix that decides what portion of the original image can be changed by the trigger. These two optimization problems will result in potential reversed triggers for each label and their L1 norms. Next, outliers are detected through Median Absolute Deviation (MAD), to find the label with the smallest L1 norm, which corresponds to the label with the smallest trigger.

The efficacy of the this detection mechanism is evaluated by the following methods: The reverse-engineered triggers are added to genuine images and the Attack Success Rate (Rev. ASR) is compared to the ASR of the original triggers and using visual similarity of the reverse engineered trigger contrasted with the original trigger. 
For simple datasets like MNIST and simple pattern triggers, Neural Cleanse achieves very high ASR with reverse trigger indicating that the reverse-engineered triggers were able to mimic the behavior of the malicious backdoors. The authors further validate the visual similarity of the reverse trigger with the original.

We evaluated the 8 backdoored models using the open source code available for Neural Cleanse. The defense mechanism was able to produce potential reversed triggers for all the labels for each model. Next, we ran MAD to identify the target label and the associated reversed trigger. Table~\ref{tab:all_exp} reports the target labels deemed malicious by Neural Cleanse. Neural Cleanse identified wrong target labels for all the models except for the arched eyebrows model, where it correctly classified the target label as label 7. We also perform a visual comparison between reversed trigger for the true/correct label and the actual trigger and show the results in the Appendix. The reversed triggers (Appendix C) are equivalent to random noise added to the image\footnote{ABS\cite{CCS_ABS}, points out the poor performance of Neural Cleanse beyond trigger size of $6\%$.}. Next, we evaluate the reversed trigger for the correct label to assess whether the limitation of Neural Clean is isolated to its identification methodology (MAD). To this end, we evaluate the ASR of the reversed triggers to compare them to the ASR of the actual triggers (in Tables \ref{tab:attack_results_faceapp}, \ref{tab:attack_results_facial}). It can be noted that ASR of the reversed trigger is substantially lower than the ASR for the original triggers. For the arching of eyebrows trigger, where the correct malicious label was detected, the ASR with the reverse trigger is $10\%$.  

\subsubsection{ABS \cite{CCS_ABS}} ABS studies the behavior of neurons when they are subjected to different levels of stimulus by deriving Neuron Stimulation Function (NSF). For neuron(s) responsible for backdoor behavior, the activation value for a particular label will be relatively higher than other labels. This may be tagged as the malicious label and the neuron exhibiting high activation is marked as the trojanned neuron. ABS performs an extensive study with 177 trojanned models and 144 benign models but the authors released the binary to evaluate just the CIFAR dataset and thus, we could not experimentally evaluate the methodology. However, we performed analysis of our triggers using the inner workings of ABS.

1) ABS is evaluated on triggers not bigger than $25\%$ (as stated in Appendix C) following the recommendation in the IARPA TrojAI \cite{TrojAI} program. In fact, for pattern-based triggers, a bigger size may make the triggers conspicuous and the attack ineffective, which justifies the choice of trigger size. Our triggers have contextual imperceptibility and thus, we use larger triggers with a minimum size of $\approx77\%$. 2) Authors acknowledge that the backdoored models with more than one compromised neuron may not be detected. Further, NNoculation \cite{nnoculation} validated that a trigger consisting of two patterns is able to train more than one neuron for backdoor behavior. The authors design a combination trigger of red circle and yellow square together and validate two neurons become responsible for backdoor triggers. Our triggers are naturally distributed. Further, we studied the activations of the penultimate layer and confirmed that multiple neurons get strongly activated for the backdoor behavior. Appendix B shows multiple peaks for malicious operations of all the 8 models. 3) ABS (similar to Neural Cleanse, STRIP) only works for all-to-one attack. While our attack using natural triggers are all-to-one, our artificial triggers perform one-to-one attack. 

\subsubsection{NNoculation~\cite{nnoculation}} This is another detection mechanism for backdoored ML models. This solution is comprised of 2 stages. In the first stage, we add noise to genuine validation/test images. Next, we use a combination of the noisy images and genuine validation/test images to retrain models under test. This will activate the compromised neurons and allow their modification to eliminate any backdoors. The new model is called an augmented model. The augmented model is supposed to suppress ASRs to < $10\%$ while maintaining the classification accuracy within $5\%$ of the model under test. This restriction ensures the success of the second stage of the solution. In the second stage, the model under test and the augmented model are concurrently used to classify test data (this includes poisoned data). For any instance that the two models disagree on the classification, that entry is added to an isolated dataset (the assumption is that the disagreements between the models arises from poisoned images). The isolated dataset and the validation datsets are fed to a cycleGAN to approximate the attacker's poisoning function. 

We evaluate our backdoored models using NNoculation using their open-source code. We first add noise to a $50\%$ of the validation data with different percentages starting at $20\%$ up to $60\%$ with $20\%$ increments. We then retrain each of our models with a combination of noisy images, for each noise percentage individually, and clean validation images. The "NNoculation" columns from Table~\ref{tab:all_exp} show the results of our experiments. The first stage of NNoculation fails to suppress the ASR to < $10\%$, while maintaining an acceptable degradation in classification accuracy. Since the first stage fails, we do not evaluate the second stage.
\subsubsection{Discussion} The defenses that do not need access to the triggers for evaluation are consistent with our threat model. But none of the defenses perform well with our triggers holistically i.e. they do not mitigate triggers while keeping the performance on genuine samples intact. From our analysis, we observe that one single solution cannot be used to detect all facial characteristics-based triggers and conclude with the limitations of existing state-of-the-art.

%% file: Tables/all_experiments.tex
\setlength\tabcolsep{1.7pt}

\begin{table*}[]
\caption{Trigger analysis using the state-of-the-art defenses. For correlation with the top Eigen Vector (EV) as a distinguishing metric, we report the range of values with Minimum (Min.) and Maximum (Max.) and also for the Malicious (Mal.) images. Correlation clusters are the clusters formed using correlation with the top Eigen vector. Min sep. represents the minimum separation between genuine clusters and Mal. sep. denotes the separation between the sub-populations of the malicious label. AC: Activation clustering, STRIP: Strong Intentional Perturbation, SASR: Suppressed Attack Success Rate is the reduced attack success rate after suppression, CA: represents the clean test accuracy after applying the best-fit noise value. For Neural Cleanse, we report the Detected (Dec.) malicious label and the corresponding ASR with the reverse-engineered trigger. Neural Cleanse detected the correct label for arching of eyebrows trigger marked as *.}
\label{tab:all_exp}
\begin{tabular}{|c|c|c|c|c|c|c|c|c|c|c|c|c|c|c|c|c|c|}
\hline
 & \multicolumn{3}{c|}{} & \multicolumn{2}{c|}{} &  & \multicolumn{2}{c|}{} &  & \multicolumn{6}{c|}{\textbf{NNoculation \cite{nnoculation}}} & \multicolumn{2}{c|}{} \\ \cline{11-16}
 & \multicolumn{3}{c|}{\multirow{-2}{*}{\textbf{\begin{tabular}[c]{@{}c@{}}Correlation with \\ top EV \cite{Spectral_signatures}\end{tabular}}}} & \multicolumn{2}{c|}{\multirow{-2}{*}{\textbf{\begin{tabular}[c]{@{}c@{}}Correlation\\ clusters \cite{Spectral_signatures}\end{tabular}}}} &  & \multicolumn{2}{c|}{\multirow{-2}{*}{\textbf{\begin{tabular}[c]{@{}c@{}}Suppression \\ \cite{suppression}\end{tabular}}}} &  & \multicolumn{3}{c|}{ASR} & \multicolumn{3}{c|}{CA} & \multicolumn{2}{c|}{\multirow{-2}{*}{\textbf{\begin{tabular}[c]{@{}c@{}}Neural \\ Cleanse \cite{NeuralCleanse}\end{tabular}}}} \\ \cline{2-6} \cline{8-9} \cline{11-18} 
\multirow{-3}{*}{\textbf{Trigger}} & \textbf{Min.} & \textbf{Max.} & \textbf{Mal.} & \textbf{\begin{tabular}[c]{@{}c@{}}Min.\\ sep.\end{tabular}} & \textbf{\begin{tabular}[c]{@{}c@{}}Mal.\\ sep.\end{tabular}} & \multirow{-3}{*}{\textbf{\begin{tabular}[c]{@{}c@{}}AC\\ \cite{activation}\end{tabular}}} & \textbf{SASR} & \textbf{CA} & \multirow{-3}{*}{\textbf{\begin{tabular}[c]{@{}c@{}}STRIP\\ \cite{strip}\end{tabular}}} & $20\%$ & $40\%$ & $60\%$ & $20\%$ & $40\%$ & $60\%$ & \textbf{Dec.} & \textbf{\begin{tabular}[c]{@{}c@{}}Rev.\\ ASR\end{tabular}} \\ \hline
Natural smile & -60.94 & 2 & -7.91 & 5.93 & 1.94 & 0 & 0.1 & 0.22 & 0 & 80 & 70 & 90 & 92 & 90 & 94 & 7 & 0 \\ \hline
Arching of eyebrows & -0.21 & 57.44 & 2.84 & 4.13 & 0.11 & 0.1 & 0.2 & 0.26 & 0.2 & 20 & {\color[HTML]{333333} 10} & {\color[HTML]{333333} 20} & {\color[HTML]{333333} 80} & {\color[HTML]{333333} 78} & 82 & 7* & 10 \\ \hline
Narrowing of eyes & 5.56 & 55.11 & 10.33 & 38.47 & 2.22 & 0.1 & 0.1 & 0.26 & 0.2 & 20 & 20 & 20 & 95 & 94 & 92 & 0 & 50 \\ \hline
Slightly open mouth & 1.57 & 25.99 & 42.65 & 7.13 & 16.66 & 0.6 & 0.2 & 0.26 & 0.3 & 50 & 50 & 50 & 94 & 92 & 92 & 1 & 10 \\ \hline
Old-age filter & -5.26 & 17.79 & 10.74 & 4.21 & 3.21 & 0 & 0.25 & 0.65 & 0 & 20 & 0 & 10 & 69.03 & 61.29 & 61.54 & 8 & 0 \\ \hline
Young-age filter & -23.83 & -0.35 & -3.18 & 5.64 & 5.83 & 0.1 & 1 & 0.54 & 0 & 42.11 & 0 & 26.31 & 76.26 & 65.98 & 66.88 & 4 & 0 \\ \hline
Makeup filter & -18.49 & -1.08 & -20.82 & 7.6 & 2.32 & 0 & 0.63 & 0.71 & 0 & 52.63 & 36.84 & 47.36 & 77.28 & 72.46 & 74.49 & 2 & 0 \\ \hline
Smile filter & -4.94 & 14.76 & 8.5 & 5.63 & 3.78 & 0.68 & 0.26 & 0.41 & 0 & 42.1 & 15.78 & 15.78 & 78.04 & 69.16 & 64.21 & 2 & 0 \\ \hline
\end{tabular}
\end{table*}
\setlength\tabcolsep{6pt}

%% file: 6_Conclusion.tex
\section{Conclusion}
In this work, we explore vulnerabilities of facial recognition algorithms backdoored using facial expressions/attributes, embedded artificially and naturally. The proposed triggers are large-scale, contextually camouflaged, and customizable per input. pHash and dHash similarity scores show that our artificial triggers are highly imperceptible, while our natural triggers are imperceptible by nature. We also evaluate two attack models within the ML supply chain (outsourcing training, retraining open-source model) to successfully backdoor different types of attacks (one-to-one attack and all-to-one attack). Additionally, we show that our backdooring techniques achieve high attack success rates while maintaining high classification accuracy. Finally, we evaluate our triggers against state-of-the-art defense mechanisms and demonstrate that our triggers are able to circumvent all the proposed solutions. Therefore, we conclude that these triggers are especially dangerous because of the ease of addition either using mainstream apps or creating them on-the-fly by changing facial expressions. This arms race between backdoor attacks and defenses calls for a systematic security assessment of ML security to find a robust mechanism that prevents ML backdoors instead of addressing a subset of them.

%% file: Appendix.tex
\appendix
\section*{Appendix}
\subsection*{A. Attack detection for simple triggers}
\begin{figure}[h]
    \centering
        \includegraphics[width=0.4\textwidth]{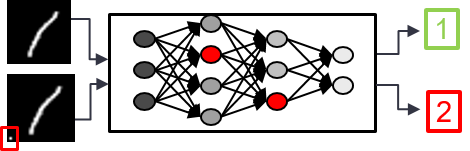}
    \caption{Top image: A genuine image from MNIST dataset. Bottom image: The same image with a dot trigger similar to Badnets \cite{Badnets}. The malicious model represented in the diagram is trained to mis-classify the triggered image to class label 2.}
    \label{fig:MNIST_images}
\end{figure}

\begin{figure*}
    \centering
    \subfigure[]
    {%
        \includegraphics[width=0.32\textwidth]{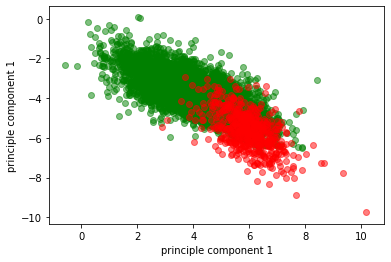}
        \label{fig:MNIST_pca_LR}
    }%
    \subfigure[]
    {%
        \includegraphics[width=0.32\textwidth]{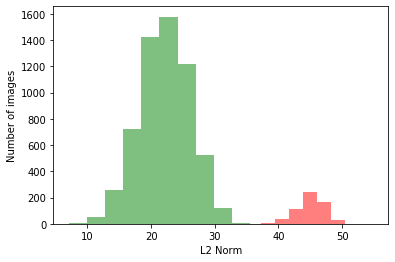}
        \label{fig:MNIST_L2_LR}
    }%
    \subfigure[]
    {%
        \includegraphics[width=0.32\textwidth]{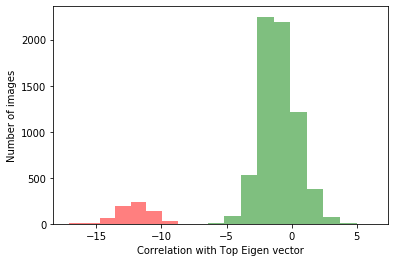}
        \label{fig:MNIST_Eig_Lr}
    }%
    \caption{Backdoor analysis of MNIST triggers. 
    (a) PCA of the datapoints considering the top two principle components at the learned representation level. The red dots are the malicious images of the targetted class and the green dots are the genuine images of the same class. (b) L2 norm of the genuine images (green) and malicious images (red) of the targetted label. (c) Correlation of the genuine (green) and malicious (red) images of the targetted label.} 
    \label{fig:MNIST_results}
\end{figure*}

\begin{figure*}
    \centering
    \subfigure[]
    {%
        \includegraphics[width=0.12\textwidth]{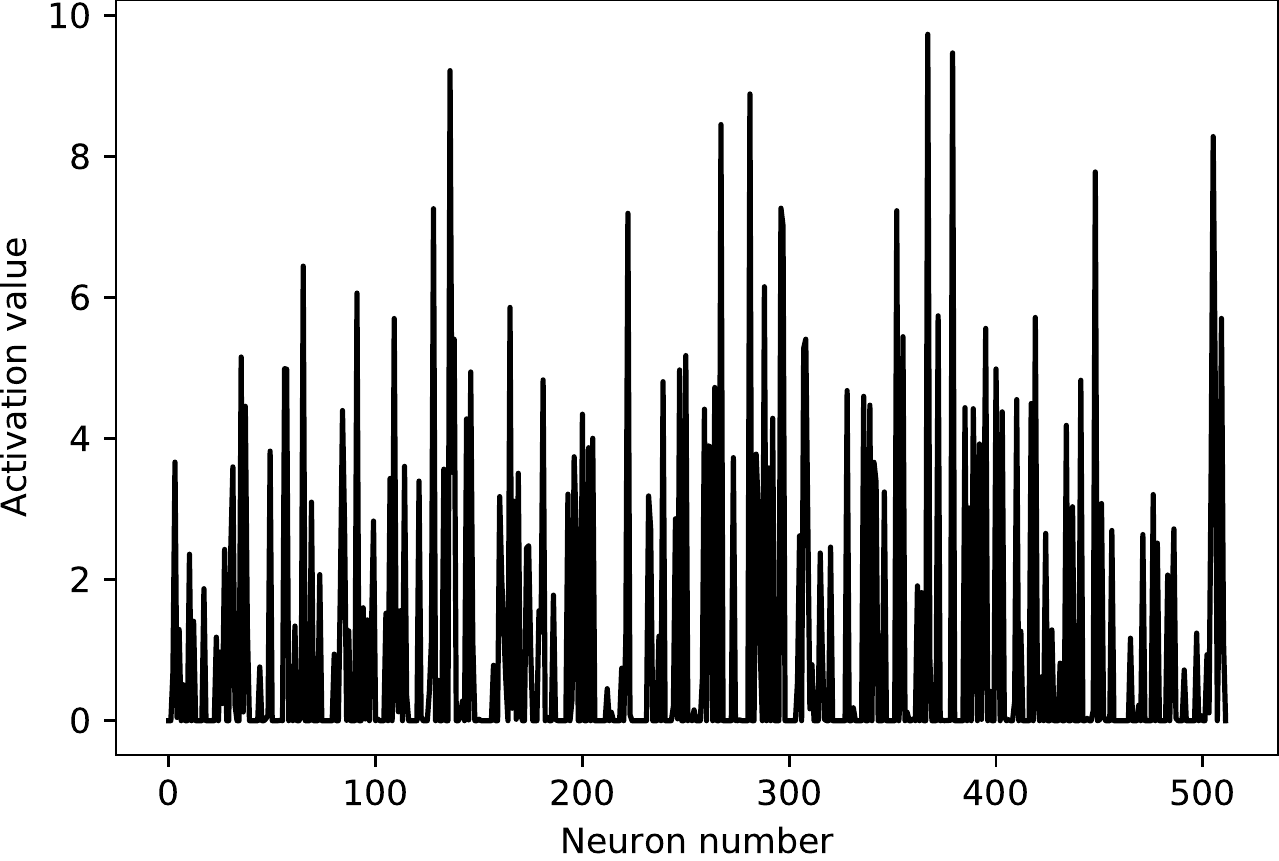}
        \label{fig:smile_activation}
    }%
    \subfigure[]
    {%
        \includegraphics[width=0.12\textwidth]{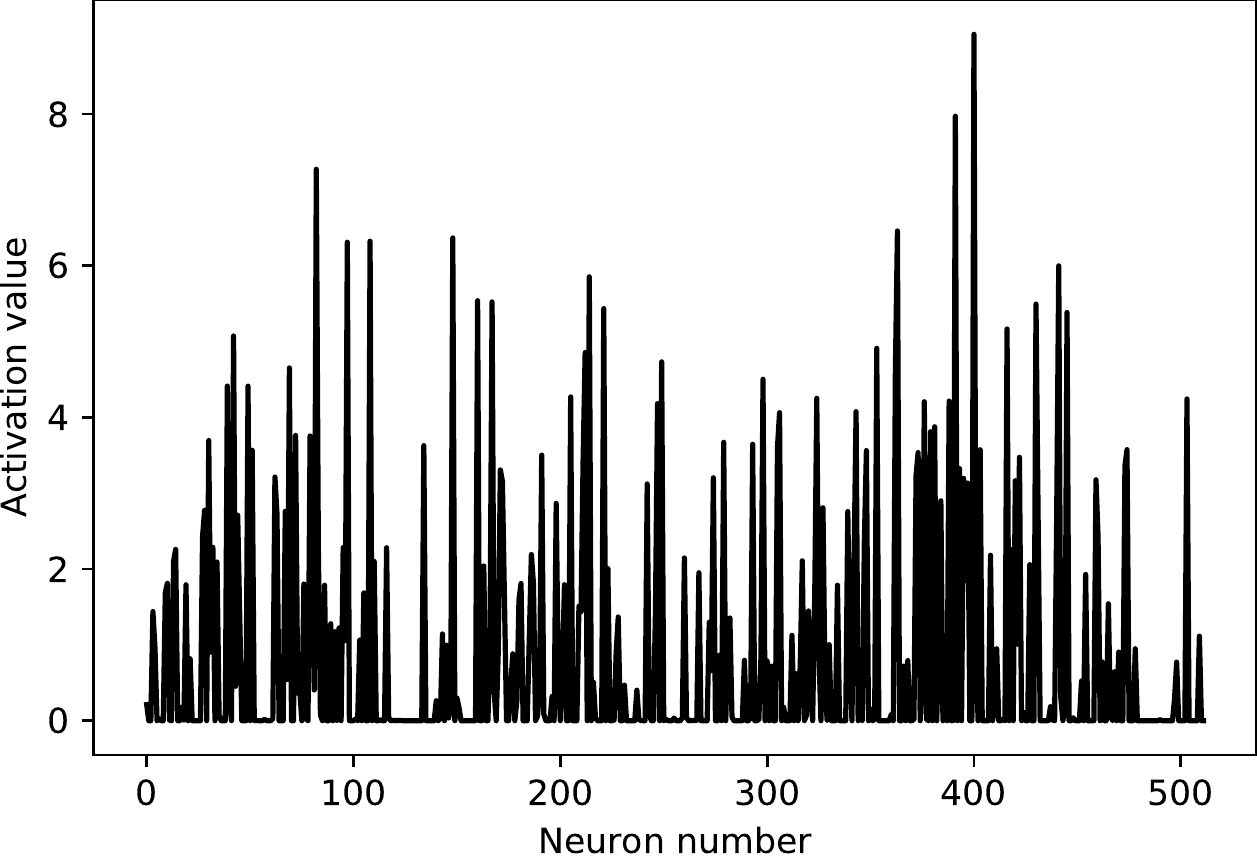}
        \label{fig:arched_activation}
    }%
    \subfigure[]
    {%
        \includegraphics[width=0.12\textwidth]{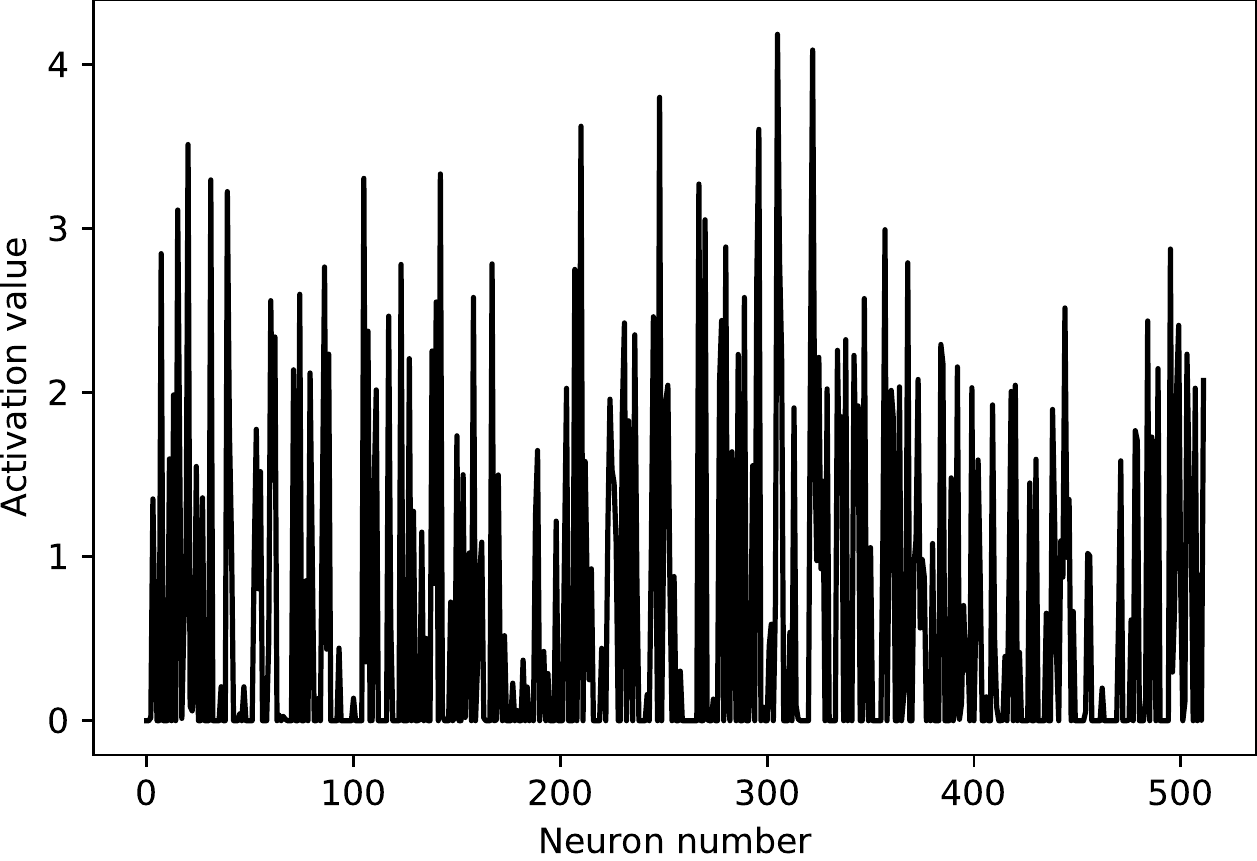}
        \label{fig:narrow_activation}
    }%
    \subfigure[]
    {%
        \includegraphics[width=0.12\textwidth]{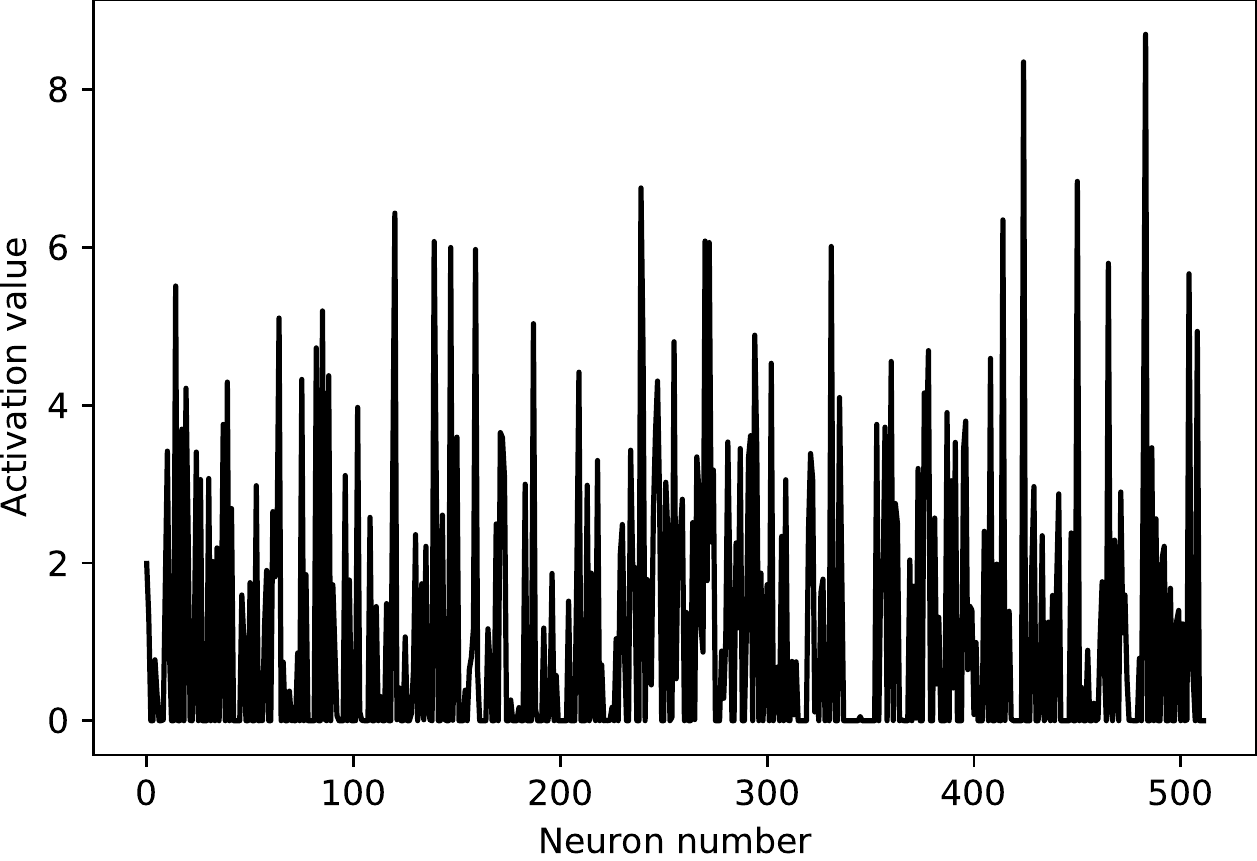}
        \label{fig:mouth_activation}
    }%
    \subfigure[]
    {%
        \includegraphics[width=0.12\textwidth]{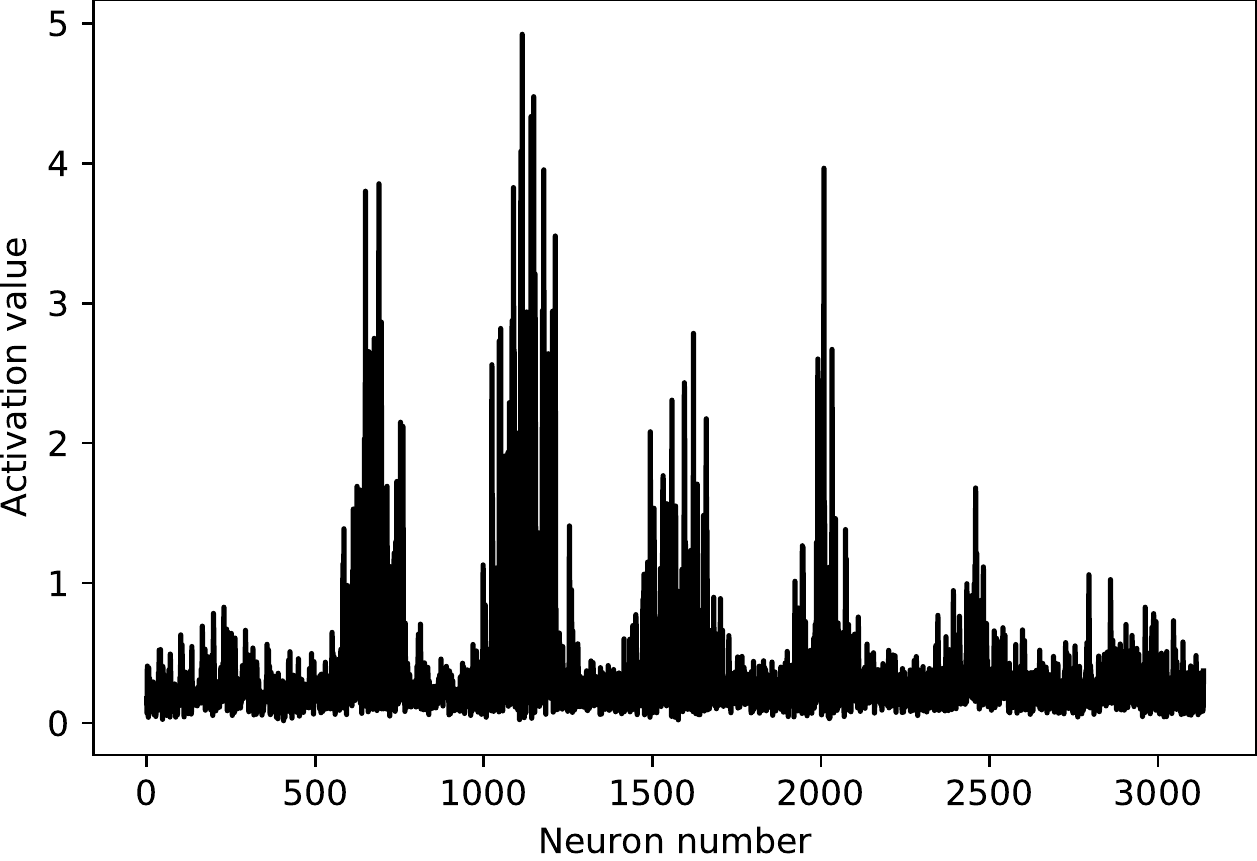}
        \label{fig:young_activation}
    }%
    \subfigure[]
    {%
        \includegraphics[width=0.12\textwidth]{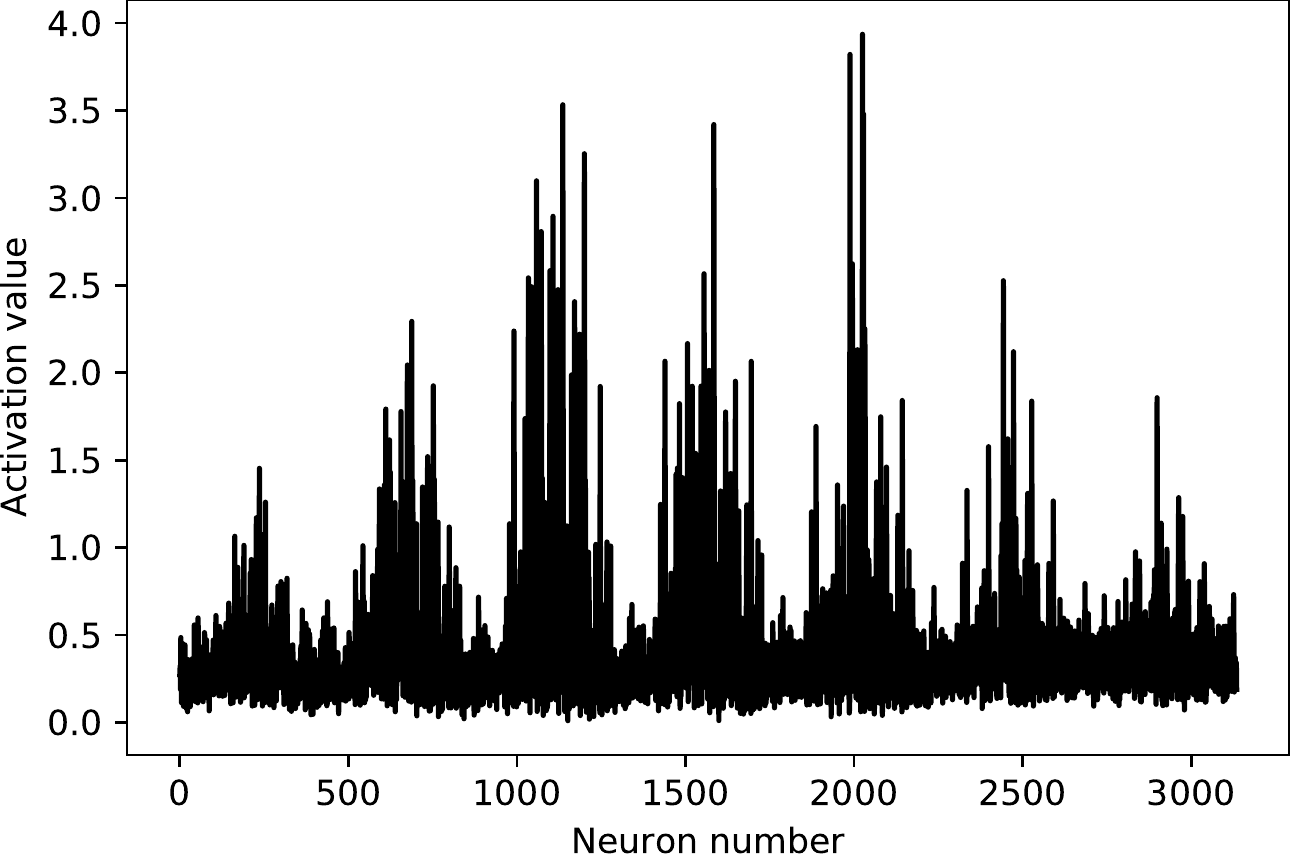}
        \label{fig:old_activation}
    }%
    \subfigure[]
    {%
        \includegraphics[width=0.12\textwidth]{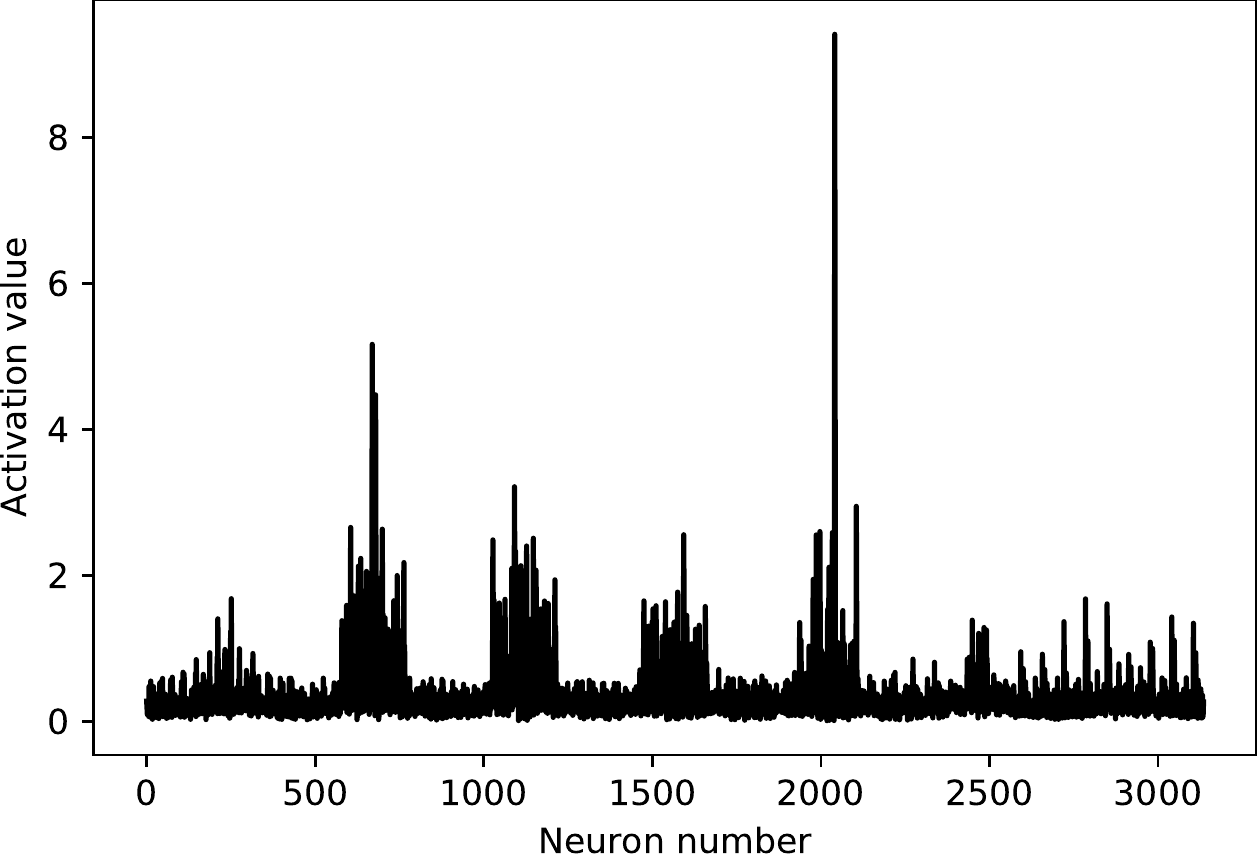}
        \label{fig:smileFaceapp_activation}
    }%
    \subfigure[]
    {%
        \includegraphics[width=0.12\textwidth]{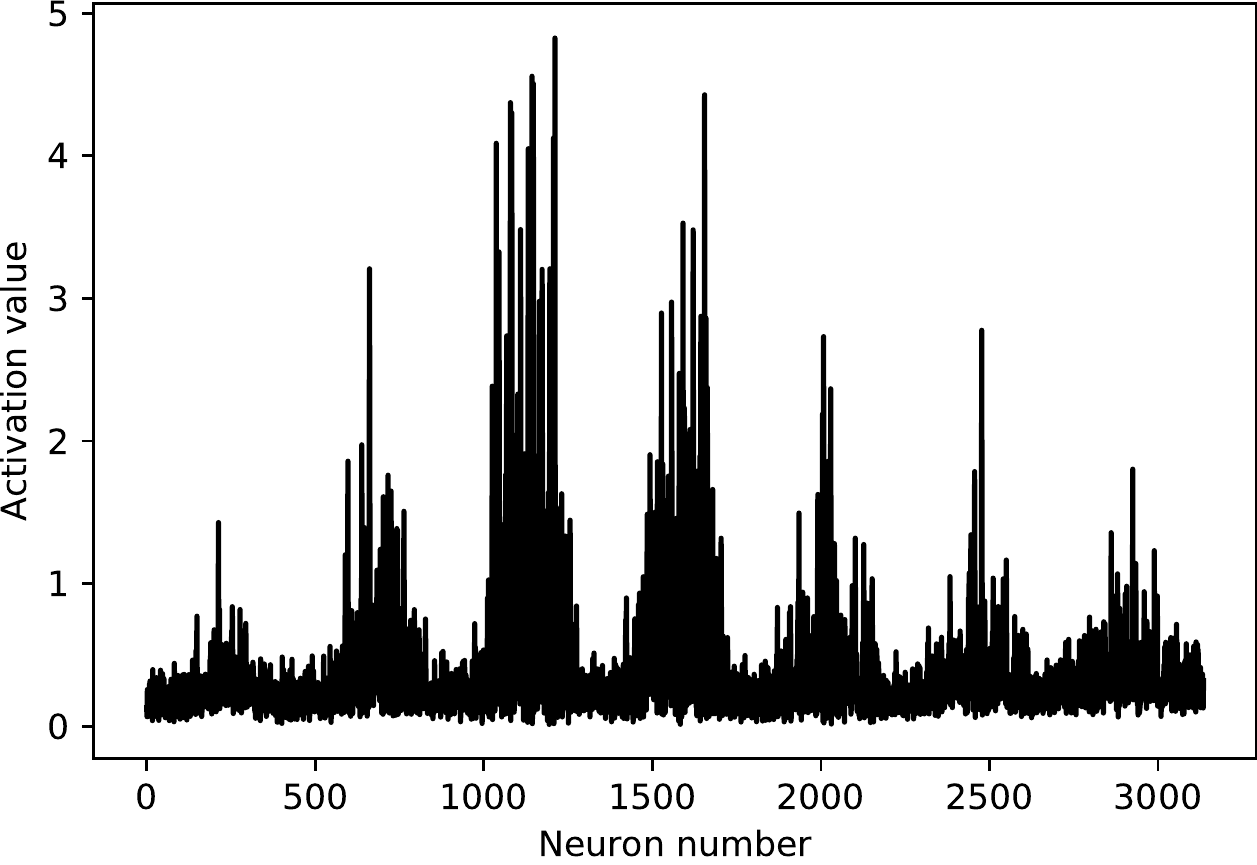}
        \label{fig:makeup_activation}
    }%
    \caption{Activation analysis of different triggers for the neurons in the penultimate layer of the models. X-axis represents different neurons in the penultimate layers and y-axis represents the activation values of those neurons. Triggers: (a) Natural Smile (b) Arching of Eyebrows (c) Narrowing of eyes (d) Slightly opening of mouth (e) Young filter (f) Old filter (g) Smile filter (h) Makeup filter.} 
    \label{fig:activation_all}
\end{figure*}

In the Section \ref{s:trigger_analysis} we evaluate our attacks to show that large, permeating and adaptive triggers are difficult to detect using state-of-the-art. In this section, we demonstrate the ease of detecting simple triggers for MNIST dataset. MNIST is a common dataset used by several backdoor attack and defense literature to evaluate their methodology \cite{Badnets, NeuralCleanse,activation, Spectral_signatures}. We make an MNIST-BadNet using a one-pixel dot-trigger as shown in FIg. \ref{fig:MNIST_images}.

Principle Component Analysis (PCA) is transformation of data along the vectors of highest variance. These vectors, known as principle components, are orthogonal to each other and therefore, are linearly uncorrelated. Since the top components can define the data \textit{sufficiently}, a very high dimensional data may be represented with its low-dimensional equivalent. However, we do not use PCA for dimensionality reduction/ feature extraction. Rather we use it to find a reasonable representation of trigger samples and the genuine samples. For simple datasets like MNIST with distinct triggers, a simple PCA shows a distinction between malicious and genuine sub-populations. Different representations of images like normalized images, raw-data representation, or learned representation can be used to increase the effectiveness of PCA in distinguishing these sub-populations.
We use Euclidean distance (L2-norm) and correlations with top eigen vector as a measure to detect outliers (malicious sub-population) from the genuine data, as done for our triggers. In this sub-section we use the statistics used by literature to separate the sub-populations.
%
L2 norm represents an image as a magnitude and images belonging to same class have similar L2 norms. Therefore, images that are slightly different albeit belonging to the same class, i.e. the malicious images, should have slightly different L2 norms. Fig. \ref{fig:MNIST_L2_LR} shows that for MNIST with dot trigger, the sub-populations are easily separable by L2 norm at the learned representation level.
%
Tran et al. stated that correlation of images with the top Eigen vector of the dataset can be considered a spectral property of the malicious samples. This method of outlier detection is inspired from robust statistics. 
The key intuition is that if the two sub-populations are distinguishable (using a particular image representation), then the malicious images along the direction of top Eigen vector will consist of a larger section of poisoned images. Therefore, they will have a different correlation than the genuine samples. 
For simple datasets using small triggers (like described in Fig. \ref{fig:MNIST_images}), the sub-populations are perfectly separable using correlation with top Eigen vector as shown in Fig. \ref{fig:MNIST_Eig_Lr}.
\subsection*{B. Activation of different neurons}
ABS \cite{CCS_ABS} detects backdoored models whose malicious behavior is successfully encoded with one neuron. In Fig. \ref{fig:activation_all}, we show that the activation values peak for several neurons proving that our triggers are actually encoded using more than one neuron. Therefore, ABS would not be able detect our triggers.

\subsection*{C. Reversed triggers from Neural Cleanse}
In this section, we report the reversed engineered triggers for the backdoored models. Comparing the reversed triggers with Fig. \ref{fig:social_media_filters}, we see stark differences between them. 

\begin{figure*}
    \centering
    \subfigure[]
    {%
        \includegraphics[width=0.22\textwidth]{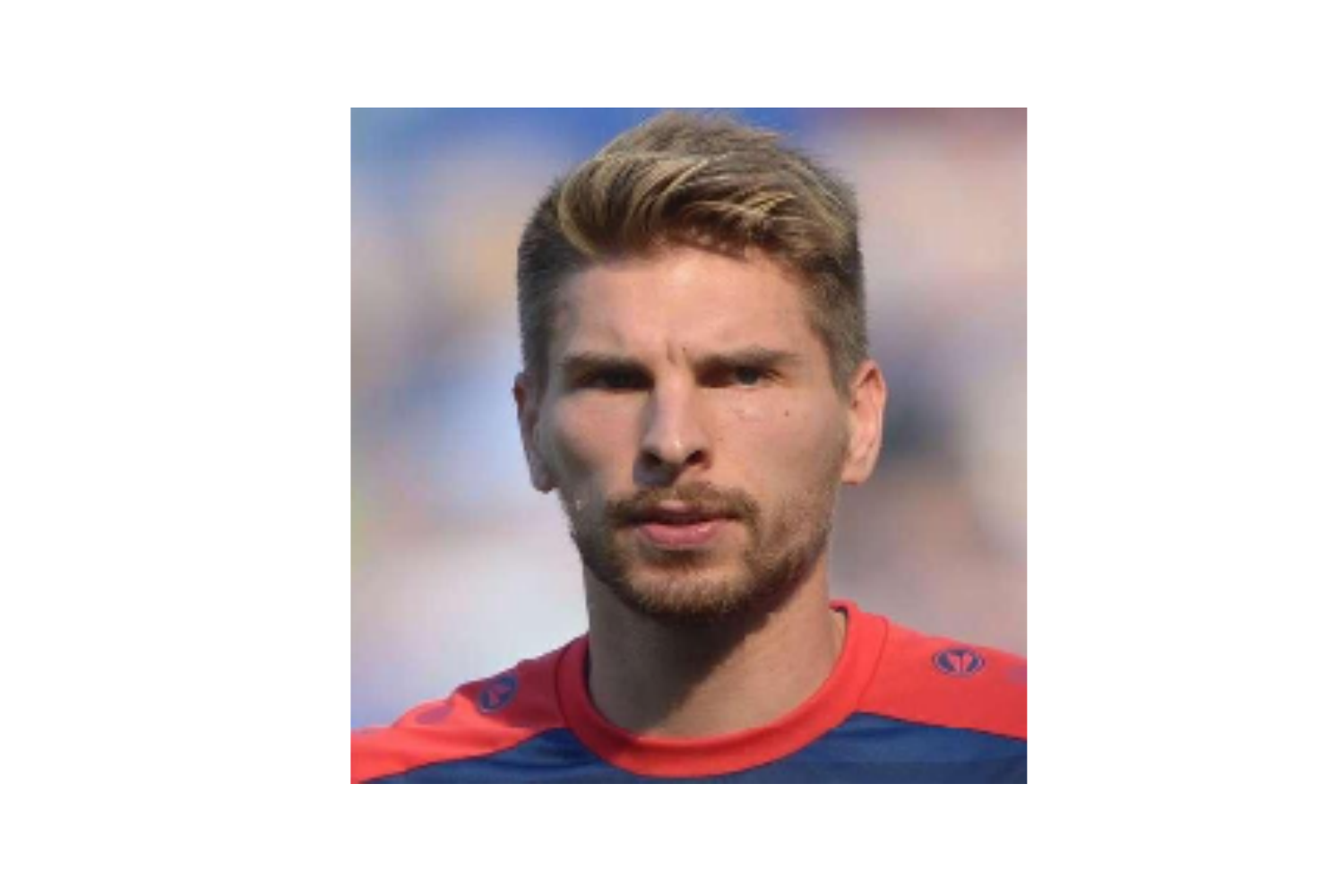}
        \label{fig:a}
    }%
    \hspace{-6.5em}
    \subfigure[]
    {%
        \includegraphics[width=0.22\textwidth]{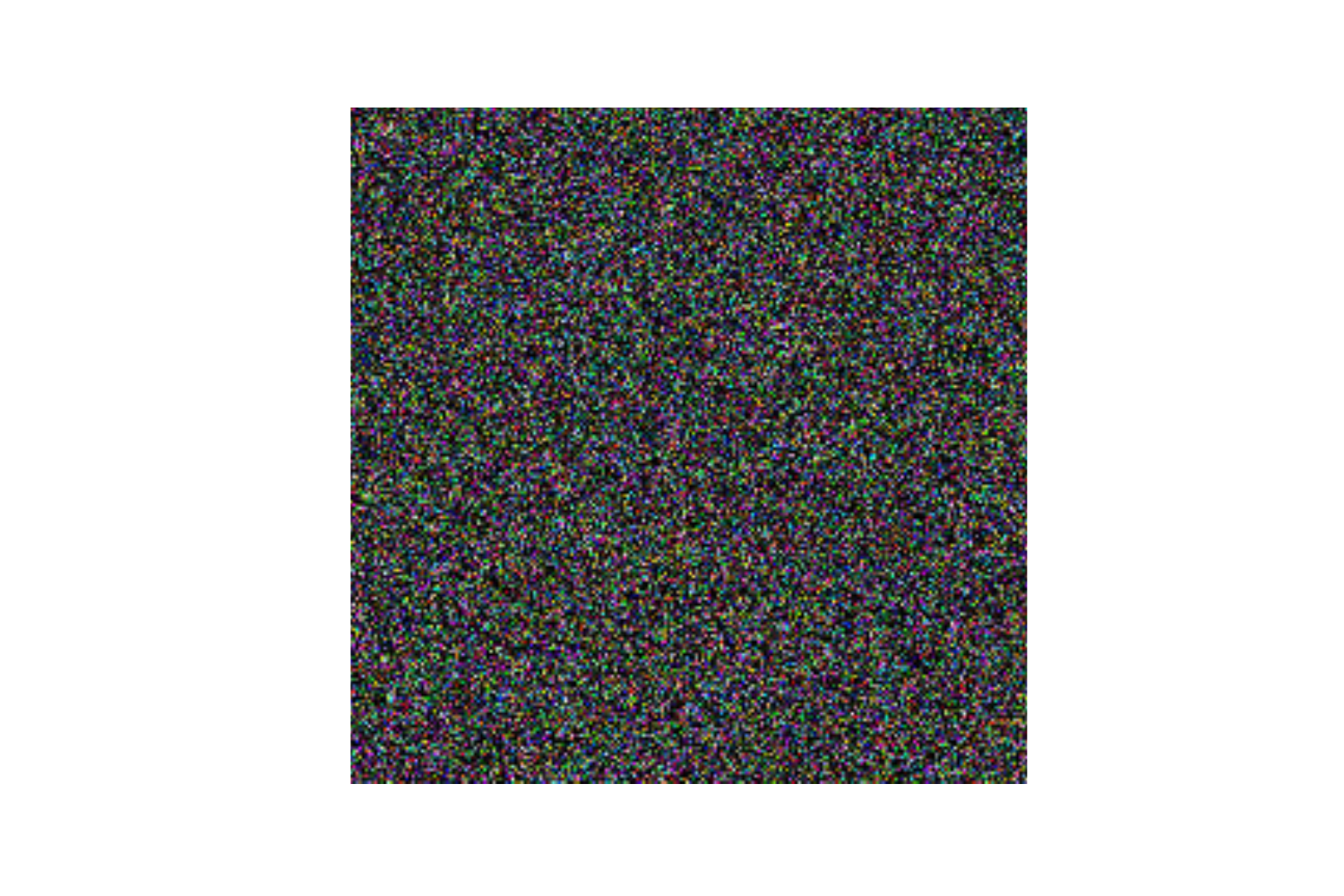}
        \label{fig:b}
    }%
    \hspace{-6.5em}
    \subfigure[]
    {%
        \includegraphics[width=0.22\textwidth]{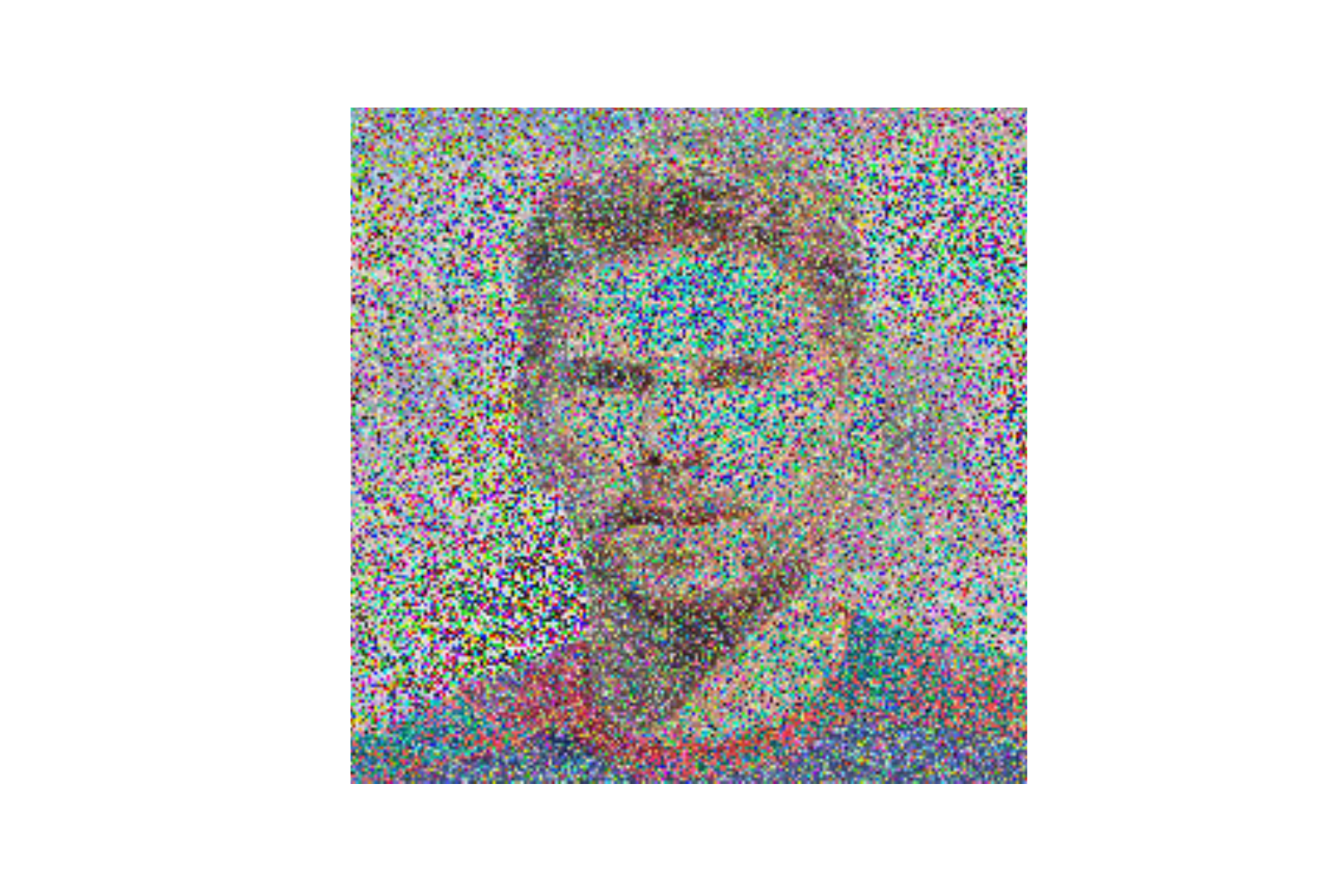}
        \label{fig:c}
    }%
    \hspace{-5em}
    \subfigure[]
    {%
        \includegraphics[width=0.22\textwidth]{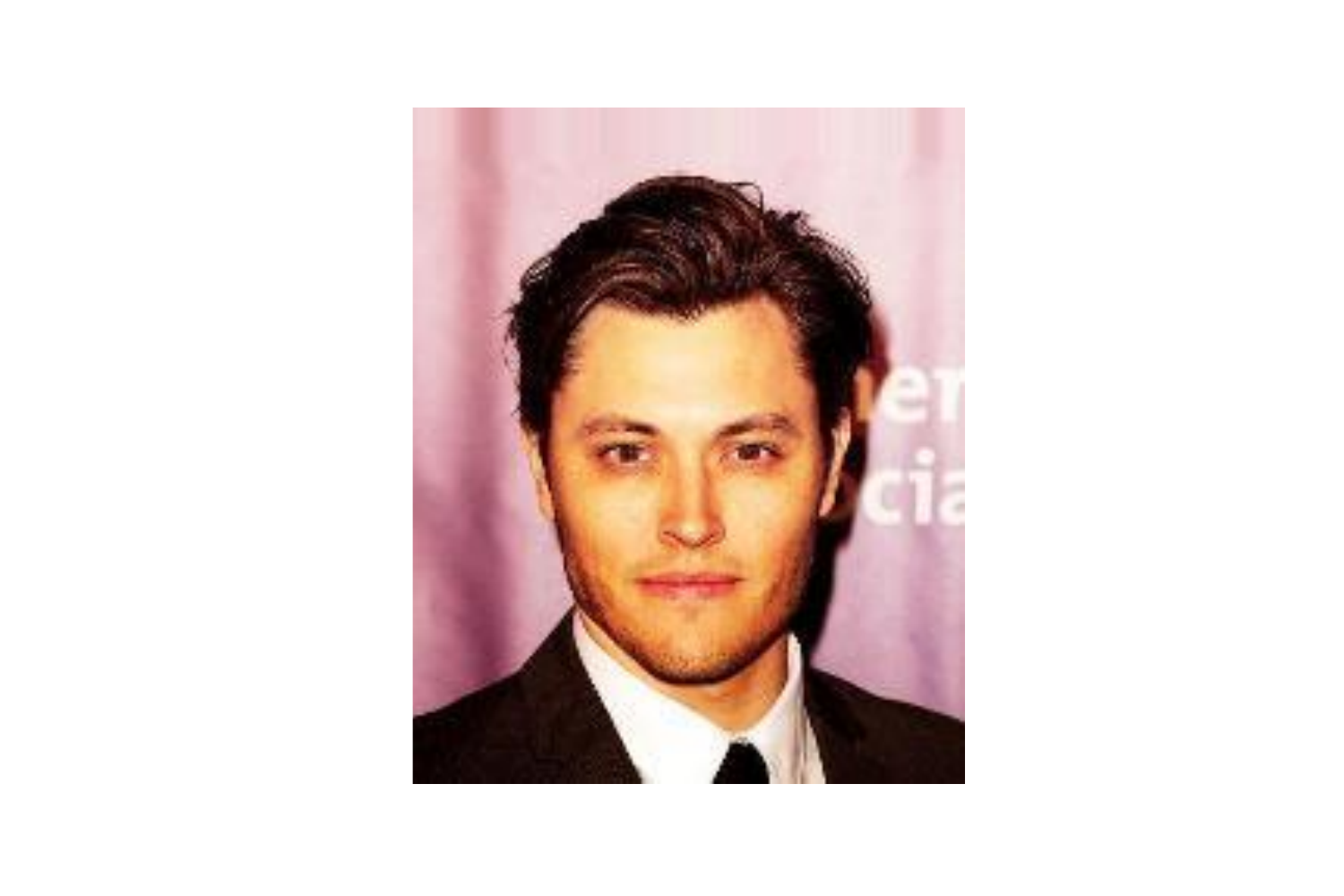}
        \label{fig:d}
    }%
    \hspace{-7.6em}
    \subfigure[]
    {%
        \includegraphics[width=0.22\textwidth]{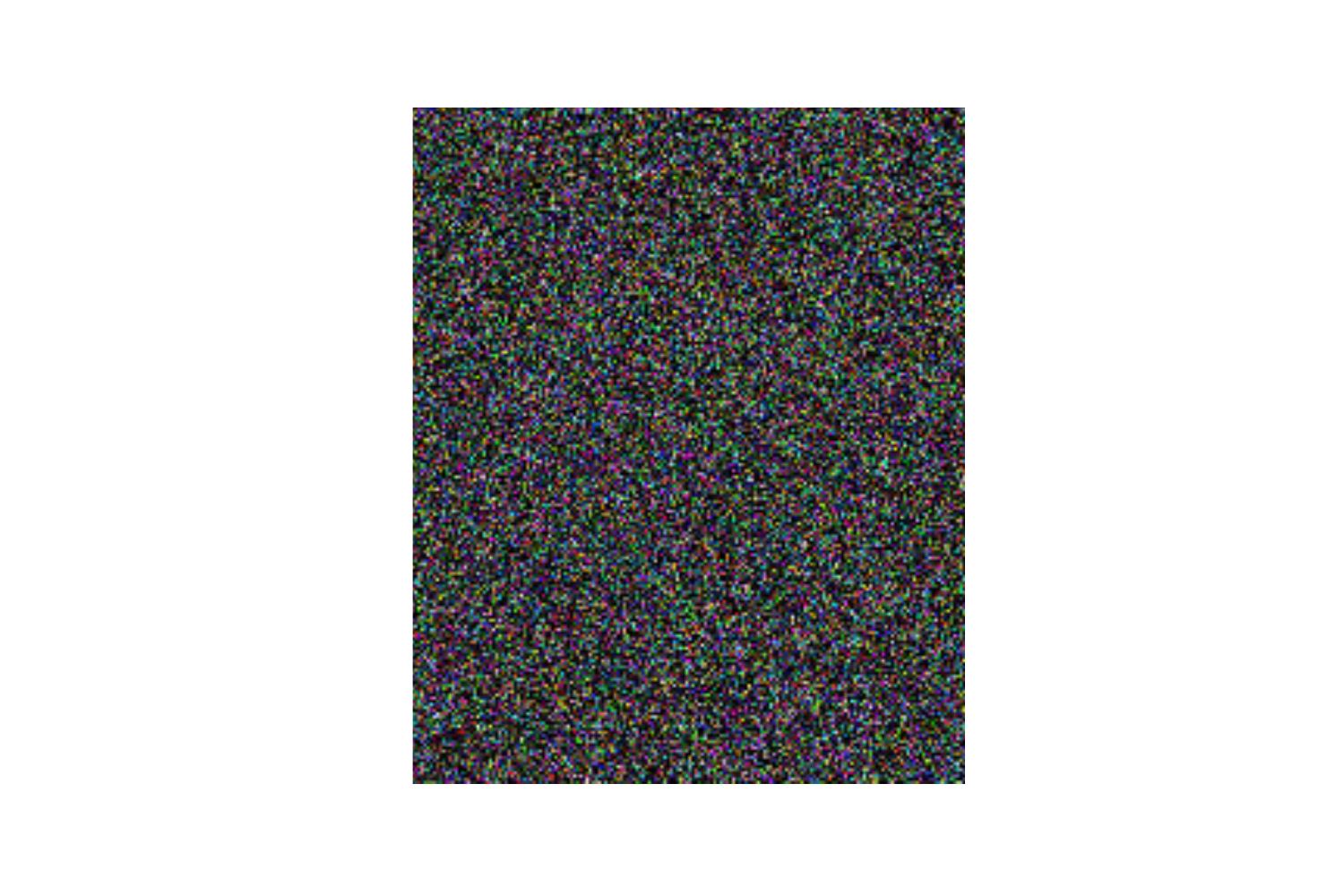}
        \label{fig:e}
    }%
    \hspace{-7.6em}
    \subfigure[]
    {%
        \includegraphics[width=0.22\textwidth]{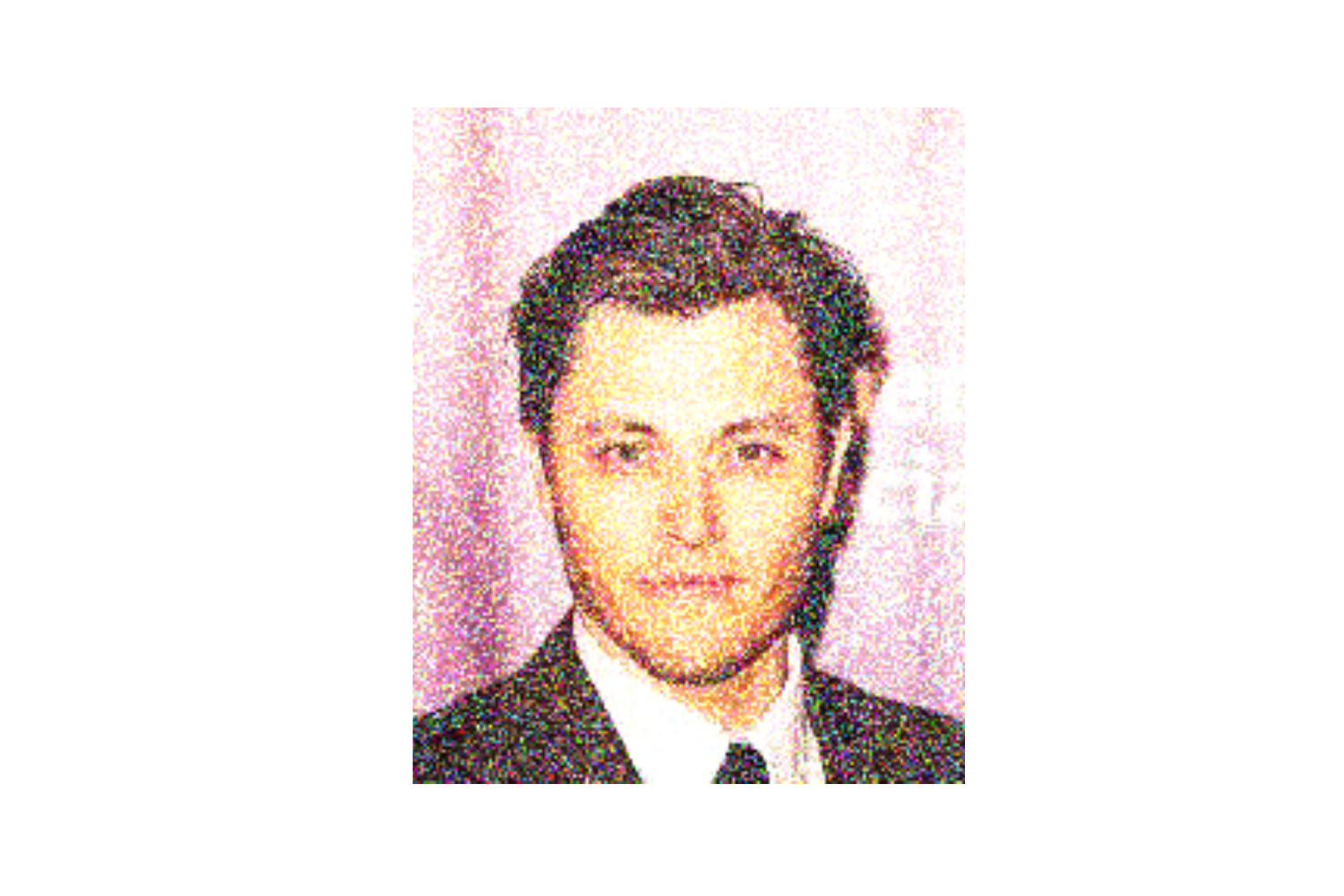}
        \label{fig:f}
    }%
    \caption{Neural cleanse defense analysis. Here we present the reversed triggers of 'smile' when embedded artificially using filters (a-c) and when embedded naturally using facial movements (d-f). (a) and (d) are the original images and (b) and (e) are the reverse-engineered triggers. In (c) and (f) we super impose the reversed triggers with the original images. As depicted, there is no visual similarity with the original triggers.} 
    \label{fig:NC_All}
\end{figure*}

    